\documentclass[runningheads]{llncs}

\usepackage{eccv}

\usepackage{eccvabbrv}

\usepackage{graphicx}
\usepackage{booktabs}

\usepackage[accsupp]{axessibility}  

\usepackage{multirow}
\usepackage{colortbl}
\usepackage{arydshln}
\usepackage[ruled,vlined]{algorithm2e}

\usepackage{hyperref}

\usepackage{orcidlink}
\usepackage{marvosym}

\begin{document}

\title{Overcome Modal Bias in Multi-modal Federated Learning via Balanced Modality Selection} 

\titlerunning{Balanced Modality Selection for MFL}

\author{Yunfeng Fan\inst{1}\orcidlink{0000-0002-2277-5355} \and
Wenchao Xu\inst{1}\textsuperscript{(\Letter)}\orcidlink{0000-0003-0983-387X} \and
Haozhao Wang\inst{2}\orcidlink{0000-0002-7591-5315} \and
Fushuo Huo\inst{1}\orcidlink{0000-0003-1030-7834} \and
Jinyu Chen\inst{1}\orcidlink{0000-0002-3502-0146} \and
Song Guo\inst{3}\orcidlink{0000-0001-9831-2202}}

\authorrunning{Y. Fan et al.}

\institute{Department of Computing, The Hong Kong Polytechnic University, Hong Kong \and
School of Computer Science and Technology, Huazhong University of Science and Technology, Hubei, China \and
Department of Computer Science and Engineering, The Hong Kong University of Science and Technology, Hong Kong\\
\email{\{yunfeng.fan,fushuo.huo,jinyu.chen\}@connect.polyu.hk}
\email{wenchao.xu@polyu.edu.hk, hz\_wang@hust.edu.cn,songguo@cse.ust.hk}
\url{https://github.com/fanyunfeng-bit/Balanced-Modality-Selection-in-MFL}}

\maketitle

\begin{abstract}
  Selecting proper clients to participate in each federated learning (FL) round is critical to effectively harness a broad range of distributed data. 
Existing client selection methods simply consider the mining of distributed uni-modal data, yet, their effectiveness may diminish in multi-modal FL (MFL) as the modality imbalance problem not only impedes the collaborative local training but also leads to a severe global modality-level bias.
We empirically reveal that local training with a certain single modality may contribute more to the global model than training with all local modalities.
%
To effectively exploit the distributed multiple modalities, we propose a novel Balanced Modality Selection framework for MFL (BMSFed) to overcome the modal bias.
On the one hand, we introduce a modal enhancement loss during local training to alleviate local imbalance based on the aggregated global prototypes.
On the other hand, we propose the modality selection aiming to select subsets of local modalities with great diversity and achieving global modal balance simultaneously. 
Our extensive experiments on audio-visual, colored-gray, and front-back datasets showcase the superiority of BMSFed over baselines and its effectiveness in multi-modal data exploitation.
  
\end{abstract}

\section{Introduction}
\label{sec:intro}

Federated learning (FL) \cite{mcmahan2017communication} aims to collaboratively learn data that has been collected by, and resides on, a number of remote devices or servers.
FL stands to develop top-performing models by aggregating knowledge from numerous edge clients \cite{dai2023tackling,wang2023aocc}, which relies on the iterative interaction among participating clients and the server.
However, comprehensively employing the information from all clients can be exceptionally difficult due to the client heterogeneity and resource limitations \cite{duan2023federated,reisizadeh2022straggler}.

Random sampling \cite{wang2023dafkd,li2019convergence} from available clients has been widely used in FL to satisfy some practical restrictions, e.g., limited communication bandwidth \cite{niknam2020federated,wang2021losp} and computing capacities \cite{imteaj2021survey}. To improve the information exploitation of all clients, extensive research has been conducted on effective client selection strategies \cite{deng2021auction,xu2020client}. 
Despite the success of traditional client selection methods in uni-modal FL, their effectiveness diminishes when dealing with clients with multi-modal data as the inter-modal interactions during the MFL training are neglected.
According to \cite{wang2020makes,huang2022modality}, there may exist inconsistent learning paces for different modalities in multi-modal joint training, i.e., \textit{modality imbalance}, which not only impedes the collaborative local training but also leads to a severe modal bias for global model in MFL. As illustrated in \cref{tab:FL modal bias}, audio modality significantly outperforms visual modality in CREMA-D and AVE datasets during local training and the aggregated model still suffers from it. 
However, two well-designed client selection methods (pow-d \cite{cho2020client} and DivFL \cite{balakrishnan2022diverse}) only obtain severely limited improvement over random sampling (FedAvg) on the multi-modal global model. 
We can see that client selection methods achieve the best uni-audio performance while uni-visual performance even drops sometimes, which means existing client selection scheme heavily relies on the better modality, while ignoring the importance of improving weak modalities that also has potential for global model aggregation.
Based on the above analysis, a pivotal question arises: \textit{Can we design a new selection scheme in MFL that can overcome the modal bias and exploit each modality comprehensively?}
\begin{table}[t]
\renewcommand\arraystretch{1.0}
\caption{Performance of various client selection methods in MFL under IID setting. A and V denote uni-audio and uni-visual while A-V means the multi-modal result. `Local' represents that a client is trained based its local data without aggregation. A strong modal bias of global model exists on the two datasets.}
\label{tab:FL modal bias}
\centering
\setlength{\tabcolsep}{3.5mm}{
\begin{tabular}{c|c|c|c|c|c|c}
\hline
  Dataset         & \multicolumn{3}{c|}{CREMA-D \cite{cao2014crema}} & \multicolumn{3}{c}{AVE \cite{tian2018audio}} \\
\hline
  Method         & A       & V       & A-V     & A      & V      & A-V   \\
\hline
Local     & 41.9    & 20.4    & 39.6    & 33.4   & 16.7   & 35.2  \\
\hline
FedAvg     & 51.2    & 20.6    & 50.7    & 61.1   & 26.8   & 62.2  \\
\hdashline
pow-d \cite{cho2020client}      & 51.5    & 20.4    & 50.5    & 61.9   & 26.9   & 62.5  \\
DivFL \cite{balakrishnan2022diverse}     & \textbf{52.3}    & 21.1    & 51.7    & \textbf{62.7}   & 25.3   & 63.3  \\
\hline
FedAvg-0.2 & 50.6    & 28.6    & 52.4    & 60.6   & 29.6   & 63.4  \\
FedAvg-0.5 & 50.5    & 34.6    & 55.7    & 58.7   & 30.0   & 60.7  \\
FedAvg-0.8 & 48.1    & \textbf{50.9}    & 61.2    & 56.4   & 31.8   & 58.5  \\ 
\hline \rowcolor{gray!20}
BMSFed     & 51.0    & 41.9    & \textbf{64.5}    & 59.7   & \textbf{40.2}   & \textbf{64.7} \\
\hline
\end{tabular}
\vspace{-10pt}
}
\end{table}

To answer this question, we investigate the interactions between different modalities via randomly discarding the data from one modality (audio or visual) on part of clients, which is inspired from modality dropout \cite{alfasly2022learnable,xiao2020audiovisual} that drops a specific modal data during training for regularization. The results are shown in rows 7-9 in \cref{tab:FL modal bias}, where `-x' denotes randomly discarding a modality on a client with probability `x'. We can see that dropping with a certain probability can improve the global multi-modal performance on both datasets (e.g., FedAvg-0.2), and the main reason comes from the dramatic improvement of visual modality. This phenomenon suggests that performing uni-modal training can unleash its potential without being inhibited by another modality, and \textbf{uni-modal local training may contribute more to the global model than multi-modal training} on some clients . 
As the dropping ratio increases, although the visual modality still improves, multi-modal performance may decline as the audio modality declines as shown in columns 5 and 7, suggesting that we should carefully control \textit{which modalities on each client should be involved in training and aggregation to make the contribution most} to the global model.
\begin{figure}[t]
    \centering
    \includegraphics[width=1.0\linewidth]{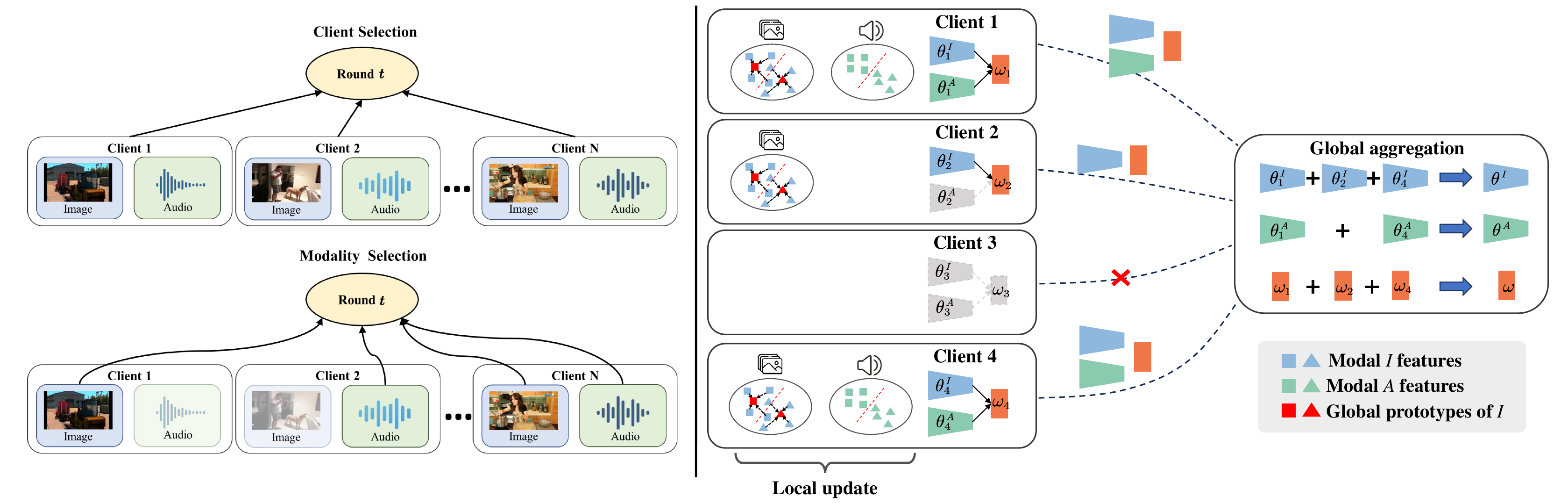}
    \caption{\textbf{Left:} Traditional client selection in FL aims to sample a client subset in each round while our modality selection considers each local modality as the sampling unit. \textbf{Right:} The paradigm of BMSFed with four clients. The global prototypes are used to enhance the weak modality during local update. Only networks corresponding to the selected modalities will be uploaded to the server for aggregation. 
    }
    \vspace{-10pt}
    \label{fig:prior and ours}
\end{figure}

According to the above investigations, we propose a novel Balanced Modality Selection scheme for MFL (BMSFed) to mitigating the modal bias and comprehensively exploit the diverse information from all modalities. Specifically, instead of selecting a subset of clients, we treat each modality on the local side as a selection unit, as demonstrated in \cref{fig:prior and ours}. Our BMSFed mainly contains two parts: Firstly, we intend to alleviate the local modality imbalance by introducing a modal enhancement loss based on aggregated global prototypes to promote the performance of weak modality. 
Secondly, we complete the modality selection by building two separated submodular functions for selecting multi-modal clients (training with multi-modal data) and uni-modal clients (training with selected uni-modal data) respectively. Inspired from \cite{balakrishnan2022diverse}, the criterion is to select modalities that are most representative on the gradients while also alleviate the global modal bias. 
A simple yet effective conflict resolution strategy is devised to ensure the validity of modality selection and keep modal balance on global model simultaneously.

The main contributions of the paper are summarized as follows:
\begin{enumerate}
\vspace{-12pt}
    \item We empirically analyze the modality imbalance problem in MFL and reveal that uni-modal training on some clients may contribute more to the global model than multi-modal training.
    \item Based on the analysis, we propose a novel Balanced Modality Selection scheme for MFL (BMSFed) to comprehensively exploit all modalities via a modal enhancement loss and representative modality selection to overcome the global modal bias.
    \item We conduct comprehensive experiments on audio-visual, colored-gray, and front-back datasets, and considering the statistical heterogeneity and modality incongruity problems in MFL, to validate the superiority of our BMSFed.
\end{enumerate}

\vspace{-5pt}
\section{Background and Related Works}
\vspace{-5pt}
\subsection{Multi-modal federated learning}
In MFL, each client has one or multiple modalities of data.
Without loss of generality, we consider two input modalities, which are denoted by $A$ and $I$ respectively in MFL.
There are a set $V$ of $N$ clients, $V=\left[ N \right] $, respectively owning datasets $\mathcal{D} _i=\left\{ \boldsymbol{X}_{i}^{A},\boldsymbol{X}_{i}^{I},\boldsymbol{y} \right\}$,
$i\in \left[ N \right]$. A typical federated learning objective is the average of each client's local loss function:
\begin{equation}
\vspace{-10pt}
    \min f\left( \theta \right) =\sum_{k=1}^N{p_kF_k\left( \theta \right)}
    \label{eq:FL loss}
\end{equation}
where $\theta =\left\{ \theta ^A,\theta ^I,\omega \right\} $ denotes the model parameters. $\theta ^A$ and $\theta ^I$ represent the encoder parameters of modality $A$ and $I$. $\omega$ is the parameter of fusion classifier. $p_k$ is a pre-defined weight.
$F_k$ is each client's local loss (cross entropy (CE) loss for classification task in this paper). 

Statistical heterogeneity \cite{karimireddy2020scaffold,li2020federateda} is a widely concerned challenge in uni-modal FL. 
To tackle this issue, FedProx \cite{li2020federatedb} uses a proximal term to stabilize model aggregation. FedProto \cite{tan2022fedproto} shares class prototypes 
to regularize the learning of local models.
In MFL, modality incongruity \cite{zhao2022multimodal,yu2023multimodal} (clients consist of different modalities combinations),
as well as statistical heterogeneity \cite{xiong2022unified}, are all considered.
Yu \etal \cite{yu2023multimodal} propose CreamFL to align the representations between different clients and different modalities via communicating knowledge on a public dataset. 
Chen \etal \cite{chen2022fedmsplit} introduce FedMSplit to split local models into several components and aggregate them by their correlations.
However, they still focus on the heterogeneity, but ignore the interaction between private data of different modalities, which limits their information exploitation.

\vspace{-5pt}
\subsection{Client selection and submodular function }
\vspace{-5pt}
Client selection \cite{deng2021auction,xu2020client} is a critical issue for FL especially when the communication cost with all devices is prohibitively high, which has been extensively studied in uni-modal FL. Cho \etal \cite{cho2020client} propose Power-of-Choice to select clients with largets local loss. Balakrishnan \etal \cite{balakrishnan2022diverse} propose to select a subset of clients with great diversity.

\noindent\textbf{Diverse client selection via submodularity.} Maximizing a submodular function is reported to improve the diversity and reduce the redundancy of a subset. This property makes it appropriate for 
client selection in FL. If a function $F$ is submodular, it should satisfy: given a finite ground set $V$ of size $N$, $F\left( A\cup \left\{ v \right\} \right) -F\left( A \right) \geqslant F\left( B\cup \left\{ v \right\} \right) -F\left( B \right)$, for any $A\subseteq B\subseteq V$ and $v\in V\backslash B$. The marginal utility of an element $v$ w.r.t. a subset $A$ is denoted as $F\left( v|A \right) =F\left( A\cup \left\{ v \right\} \right) -F\left( A \right) $, which can represent the importance of $v$ to $A$. 
The client selection via submodular maximization can be expressed following \cite{balakrishnan2022diverse}: find a subset $S$ of clients whose aggregated gradients can approximate the full aggregation from all clients:
\begin{equation}
    \begin{aligned}
    \sum_{k\in \left[ N \right]}{\nabla F_k\left( v^k \right)}=\sum_{k\in \left[ N \right]}{\left[ \nabla F_k\left( v^k \right) -\nabla F_{\sigma \left( k \right)}\left( v^{\sigma \left( k \right)} \right) \right]} 
    +\sum_{k\in S}{\gamma _k\nabla F_k\left( v^k \right)}
    \end{aligned}
    \label{eq:client selection approximation}
\end{equation}
where $\sigma$ maps $V\rightarrow S$ and the gradient $\nabla F_k\left( v^k \right)$ from client $k$ is approximated by the gradient from a selected client $\sigma \left( k \right) \in S$. For $i \in S$, let $C_i\triangleq \left\{ k\in V|\sigma \left( k \right) =i \right\} $, and therefore $\gamma _i\triangleq |C_i|$. Take the norms and apply triangular inequality after subtracting the second term from both sides, we can obtain a relaxed objective $G\left( S \right)$ for minimizing the approximation error:
\begin{equation}
    \begin{split}
        \left\| \sum_{k\in \left[ N \right]}{\nabla F_k\left( v^k \right)}-\sum_{k\in S}{\gamma _k\nabla F_k\left( v^k \right)} \right\| \leqslant 
        \sum_{k\in \left[ N \right]}{\underset{i\in S}{\min}\left\| \nabla F_k\left( v^k \right) -\nabla F_i\left( v^i \right) \right\| \triangleq G\left( S \right)}
    \end{split}
    \label{eq:relaxed objective for client selection}
\end{equation}
Minimizing $G\left( S \right)$ can be seen as maximizing the well-known submodular function, i.e., the facility location function \cite{cornuejols1977uncapacitated}. The submodular maximizing problem is NP-hard but can be approximated via the greedy \cite{nemhauser1978analysis} or stochastic greedy algorithm \cite{mirzasoleiman2015lazier}:
\begin{equation}
    \begin{aligned}
        S\gets S\cup k^*, 
        k^*\in \underset{k\in \mathrm{rand}\left( V\backslash S, \mathrm{size}=s \right)}{\mathrm{arg}\max}\left[ \bar{G}\left( S \right) -\bar{G}\left( \left\{ k \right\} \cup S \right) \right] 
    \end{aligned}
    \label{eq:stochastic greedy for client selection}
\end{equation}
$\bar{G}$ represents a constant minus the negation of $G$.

Although these methods make great improvement in uni-modal FL, the selection strategy is under-explored in MFL and we reveal that traditional client selection approaches cannot address the severe modal bias in MFL.  

\vspace{-5pt}
\subsection{Imbalanced multi-modal learning}
\label{sec:imbalanced MML}
\vspace{-5pt}

Modality imbalance indicates the inconsistent learning progress of different modalities in multi-modal learning \cite{wang2020makes,huang2022modality}. Peng \etal \cite{peng2022balanced} propose OGM-GE to alleviate the inhibitory effect on weak modality by slowing down the dominant modality. Fan \etal \cite{fan2023pmr} further build a non-parametric classifier by class centroids to adjust the update direction of weak modality. In this paper, we aim to power each modality of all clients by a meticulously designed modality selection strategy in each round of training.

\vspace{-7pt}
\section{Method}
\label{sec:method}
\vspace{-7pt}
In this section, we introduce BMSFed that contains the local imbalance alleviation and balanced modality selection. 
%

\vspace{-10pt}
\subsection{Local Imbalance Alleviation}
\label{sec:LIA}
As discussed in \cref{sec:imbalanced MML} and \cref{tab:FL modal bias}, the multi-modal training on each client may suffer from severe modality imbalance, leading to inadequate information exploitation and consequently incurring the modal bias on the global model. Therefore, we first try to alleviate the imbalance during local training.

According to \cite{fan2023pmr}, the prototypes (i.e., class centroids) are suitable to calibrate the gradient directions to avoid the interference from other modalities. Therefore, we choose to use prototypes to facilitate learning of weak modality. For the $i$-th client with the data $\mathcal{D} _i=\left\{ \boldsymbol{X}_{i}^{A},\boldsymbol{X}_{i}^{I},\boldsymbol{y} \right\}$, the local prototype for class $j$ of each modality is defined as the mean value of representations:
\begin{equation}
\begin{aligned}
  c_{i,j}^{I}=\frac{1}{\left|\mathcal{D} _{i,j}\right|}\sum_{x^I\in \mathcal{D} _{i,j}}{h_i\left( \theta _i^I;x^I \right)},
  c_{i,j}^{A}=\frac{1}{\left|\mathcal{D} _{i,j}\right|}\sum_{x^A\in \mathcal{D} _{i,j}}{h_i\left( \theta _i^A;x^A \right)} 
  \label{eq:local prototype}
\end{aligned}
\end{equation}
where $\mathcal{D} _{i,j}$ denotes the samples belonging to $j$-th class in client $i$. $h_i$ is the function of encoder. Considering the heterogeneity across clients, we aggregate the local prototypes to a global prototype as:
\begin{equation}
\begin{aligned}
  c_{j}^{GI}=\frac{1}{\left| \mathcal{N} _j \right|}\sum_{i\in \mathcal{N} _j}{\frac{\left| \mathcal{D} _{i,j} \right|}{N_j}c_{i,j}^{I}},
  c_{j}^{GA}=\frac{1}{\left| \mathcal{N} _j \right|}\sum_{i\in \mathcal{N} _j}{\frac{\left| \mathcal{D} _{i,j} \right|}{N_j}c_{i,j}^{A}} 
  \label{eq:global prototype}
\end{aligned}
\end{equation}
where $\mathcal{N} _j$ denotes the set of clients that have class $j$ and $N_j$ is the number of instances belonging to class $j$ over all clients. Then, we introduce a modal enhancement loss (ME) based on global prototype to adjust local training:
\begin{equation}
    \begin{aligned}
        \mathcal{L} _{ME}^{k}\left( v_I^k \right) &=-\mathbb{E} _{\left( x_{i}^{I},y \right) \in \mathcal{D} _k}\log \left[ \frac{\exp \left( -d\left( z_{i}^{I},c_{y}^{GI} \right) \right)}{\sum\nolimits_{j=1}^Y{\exp \left( -d\left( z_{i}^{I},c_{j}^{GI} \right) \right)}} \right] \\
        \mathcal{L} _{ME}^{k}\left( v_A^k \right) &=-\mathbb{E} _{\left( x_{i}^{A},y \right) \in \mathcal{D} _k}\log \left[ \frac{\exp \left( -d\left( z_{i}^{A},c_{y}^{GA} \right) \right)}{\sum\nolimits_{j=1}^Y{\exp \left( -d\left( z_{i}^{A},c_{j}^{GA} \right) \right)}} \right] 
    \end{aligned}
    \label{eq:global PCE}
\end{equation}
where $d\left( \cdot ,\cdot \right)$ is the distance function (Euclidean distance), $z_{i}^{I}$ is the representation of $x_{i}^{I}$, i.e., $z_{i}^{I}=h_i\left( \theta _i^I;x_i^I \right)$. $Y$ is the class number. $v_{A}^{k}$ and $v_{I}^{k}$ indicate the corresponding modal data in client $k$. Hence, the local loss should be (data superscripts are omitted for simplicity):
\begin{equation}
    \begin{aligned}
        F_k\left( v_A,v_I \right) =\left\{ \begin{array}{c}
	\begin{matrix}
	\mathcal{L} _{CE}^{k}\left( v_{A}^{},v_{I}^{} \right) +\gamma ^k\mathcal{L} _{ME}^{k}\left( v_{A}^{} \right),&		\rho _{I}^{k}\leqslant 1\\
\end{matrix}\\
	\begin{matrix}
	\mathcal{L} _{CE}^{k}\left( v_{A}^{},v_{I}^{} \right) +\beta ^k\mathcal{L} _{ME}^{k}\left( v_{I}^{} \right),&		\rho _{I}^{k}>1\\
\end{matrix}\\
\end{array} \right. 
\label{eq:local loss}
    \end{aligned}
\end{equation}
where $\rho _{I}^{k}$ indicates the local imbalance ratio for client $k$ calculated based on the modality-wise ground truth prediction, and $\rho _{I}^{k}>1$ means modality $A$ outperforms modality $I$ and vice verse. $\gamma ^k,\beta ^k\in \left( 0,1 \right) $ are the modulation coefficients. The calculation details for $\gamma ^k$, $\beta^k$ and $\rho_I^k$ are in Appendix.

Different from \cite{fan2023pmr}, we apply the global prototype instead of local prototypes to aggregate the knowledge from different clients, and the ME loss is not only applied to multi-modal clients, but also to uni-modal clients (See details in \cref{sec:BMS}). Now, the weak modality will be stimulated through local update.

\vspace{-10pt}
\subsection{Balanced Modality Selection}
\label{sec:BMS}
\vspace{-5pt}

Considering that different types of modal combinations may be selected to participate in local training in our scheme, we define the participated clients as multi-modal clients and uni-modal clients (e.g., client 2 in \cref{fig:prior and ours} Right is a uni-modal client with modal $I$ for training while client 1 and 4 are multi-modal clients with both modalities for training). %
In \cref{sec:LIA}, we propose the ME loss to facilitate the local learning of weak modality, which is originally designed for multi-modal clients. Here, we also apply it on uni-weak-modal clients to realize gradient consistency on different types of clients. 

Assume modality $I$ is weak here. Hence, the local loss for multi-modal and uni-modal clients should be:
\begin{equation}
\begin{aligned}
\mathrm{multi}\mbox{-}\mathrm{modal}&: F_k\left( v_{A}^{},v_{I}^{} \right) =\mathcal{L} _{CE}^{k}\left( v_{A}^{},v_{I}^{} \right)+\beta ^k\mathcal{L} _{ME}^{k}\left( v_{I}^{} \right)
\\
\mathrm{uni}\mbox{-}\mathrm{modal}&: F_k\left( v_{A}^{} \right) =\mathcal{L} _{CE}^{k}\left( v_{A}^{} \right), F_k\left( v_{I}^{} \right) =\mathcal{L} _{CE}^{k}\left( v_{I}^{} \right) +\beta ^k\mathcal{L} _{ME}^{k}\left( v_{I}^{} \right)
\end{aligned}
\label{eq:multiuni local loss}
\end{equation}


Next, following \cref{eq:client selection approximation}, we define the paradigm of modality selection for MFL, aiming to approximate the full gradient aggregation from all clients by the gradient from the customized multi-modal and uni-modal clients:
\begin{equation}
    \begin{aligned}
&\sum_{k\in \left[ N \right]}{\nabla F_k\left( v_{A}^{},v_{I}^{} \right)}=\sum_{k\in \left[ N \right]}{\left[ \begin{array}{c}
	\nabla F_k\left( v_{A}^{},v_{I}^{} \right) -\nabla F_{\sigma _M\left( k \right)}\left( v_{A}^{},v_{I}^{} \right)\\
	-\nabla F_{\sigma _A\left( k \right)}\left( v_{A}^{} \right) -\nabla F_{\sigma _I\left( k \right)}\left( v_{I}^{} \right)\\
\end{array} \right]}
\\
&+\sum_{k\in S_M}{\gamma _{k}^{M}\nabla F_k\left( v_{A}^{},v_{I}^{} \right)}+\sum_{k\in S_A}{\gamma _{k}^{A}\nabla F_k\left( v_{A}^{} \right)}+\sum_{k\in S_I}{\gamma _{k}^{I}\nabla F_k\left( v_{I}^{} \right)}
    \end{aligned}
    \label{eq:modality selection}
\end{equation}
where $\sigma _M $, $\sigma _A$ and $\sigma _I$ map $V\rightarrow S_M, S_A, S_I$, the client sets who use multi-modal, uni-$A$, and uni-$I$ data for training respectively, and $S_M\cap S_A=S_A\cap S_I=S_M\cap S_I=\oslash $. 
$\gamma _k^M$, $\gamma _k^A$ and $\gamma _k^I$ have the similar meaning as $\gamma _k$ in \cref{eq:client selection approximation}.

Since modality $I$ is weak here, we omit the uni-$A$ clients as the multi-modal gradient is dominated by modality $A$ \cite{peng2022balanced}, which means we do not need to select uni-$A$ clients for its enhancement. Then, bring \cref{eq:multiuni local loss} to \cref{eq:modality selection} and follow the operations from \cref{eq:client selection approximation} to \cref{eq:relaxed objective for client selection}, we can obtain:
\begin{equation}
    \begin{aligned}
&\sum_{k\in \left[ N \right]}{\underset{i\in S_M,j\in S_I}{\min}\left\| \nabla F_k\left( v_{A}^{},v_{I}^{} \right) -{\gamma _i^M}\nabla F_i\left( v_{A}^{},v_{I}^{} \right) -{\gamma _j^I}\nabla F_j\left( v_{I}^{} \right) \right\|}
\\
&=\sum_{k\in \left[ N \right]}{\underset{i\in S_M,j\in S_I}{\min}\left\| \begin{array}{c}
	\nabla \mathcal{L} _{CE}^{k}\left( v_{A}^{},v_{I}^{} \right) +\nabla \beta ^k\mathcal{L} _{ME}^{k}\left( v_{I}^{} \right) -\nabla \mathcal{L} _{CE}^{i}\left( v_{A}^{},v_{I}^{} \right)\\
	-\nabla \beta ^i\mathcal{L} _{ME}^{i}\left( v_{I}^{} \right) -\nabla \mathcal{L} _{CE}^{j}\left( v_{I}^{} \right) -\nabla \beta ^j\mathcal{L} _{ME}^{j}\left( v_{I}^{} \right)\\
\end{array} \right\|}
\\
&\leqslant \sum_{k\in \left[ N \right]}{\underset{i\in S_M}{\min}\left\| \nabla \mathcal{L} _{CE}^{k}\left( v_{A}^{},v_{I}^{} \right) -\nabla \mathcal{L} _{CE}^{i}\left( v_{A}^{},v_{I}^{} \right) \right\|}
\\
&+\sum_{k\in \left[ N \right]}{\underset{i\in S_M,j\in S_I}{\min}\left\| \begin{array}{c}
	\nabla \beta ^k\mathcal{L} _{ME}^{k}\left( v_{I}^{} \right) -\nabla \beta ^i\mathcal{L} _{ME}^{i}\left( v_{I}^{} \right)\\
	-\nabla \mathcal{L} _{CE}^{j}\left( v_{I}^{} \right) -\nabla \beta ^j\mathcal{L} _{ME}^{j}\left( v_{I}^{} \right)\\
\end{array} \right\|}
\\
&\triangleq G\left( S_M \right) +G\left( S_M\cup S_I \right) 
    \end{aligned}
    \label{eq:modality selection decoupling}
\end{equation}

The right-hand side of the first equation is the modality selection formula that aims to select a group of multi-modal clients $S_M$ and uni-modal clients $S_I$ to approximate the aggregated gradients from all clients. However, the joint selection for $S_M$ and $S_I$ is a complex joint optimization problem. Therefore, we decouple this objective into two submodular functions $G\left( S_M \right)$ and $G\left( S_M\cup S_I \right)$ according to triangle inequality, while the full gradient approximation is divided into two parts: the first part uses selected multi-modal CE gradient to fit fully multi-modal CE gradient aggregation, and the second part approximates the fully multi-modal ME gradient aggregation via selected uni-modal CE gradient and both selected multi- and uni-modal ME gradient. The modality-level gradient decoupling converts the complex joint selection to two simply separated selection problems.

\begin{algorithm}[t]
\caption{BMSFed.}
\label{alg:BMSFed}
\KwIn{Input data $\mathcal{D} _i=\left\{ \boldsymbol{X}_{i}^{A},\boldsymbol{X}_{i}^{I},y \right\}$, $i\in \left[ N \right]$, initial model $\theta$, hyper-parameters $\chi$, global communication epochs $E$, $e=1$.} 
\BlankLine
\While{\textnormal{$e < E$}}{
    \eIf{$e = 1$}{
        Send $\theta$ to all clients; \\
        Perform one-step local update for gradients, prototypes and $\rho_I^k$; \\
        Aggregate global prototypes and $\rho_I$; \\
}
      {Aggregate global model $\theta$, prototypes $c^G$ and $\rho_I$;}
      Select multi-modality for $S_M$ and uni-modality for $S_I$ (or $S_A$)
using \cref{eq:select multi-client,eq:select uni-client}; \\
    Send $\theta$, $c^G$ and $\rho_I$ to selected clients;\\
    \ForEach{\textnormal{client in selected clients \textbf{in parallel}}}{
        Perform multi-modal learning in $S_M$ and uni-modal learning in $S_I$ (or $S_A$) by with \cref{eq:local loss,eq:multiuni local loss}; \\
        Send gradients, local prototypes and $\rho_I^k$ to server;
    }     
}
\end{algorithm}


Although we can solve the two submodular functions with the stochastic greedy algorithm \cite{mirzasoleiman2015lazier}, there are still two issues: (1) the selected client according to $G\left( S_M\cup S_I \right)$ should be specified whether it is uni-modal client or multi-modal client; (2) the separated selection strategy pays less attention to the global modal bias. To address the two problems, we perform the stochastic greedy algorithm for two submodular functions in parallel and propose a simple yet effective conflict resolution strategy to ensure $S_M\cap S_I=\oslash $ as well as, more importantly, balance the learning of different modalities on global model: 
\begin{equation}
    \begin{aligned}
        S_M\gets S_M\cup k_{1}^{*},   k_{1}^{*}\in \underset{k\in \mathrm{rand}\left( V\backslash S_M\backslash S_I,\mathrm{s} \right)}{\mathrm{arg}\max}\left[ \bar{G}\left( S_M \right) -\bar{G}\left( \left\{ k \right\} \cup S_M \right) \right] 
    \end{aligned}
    \label{eq:select multi-client}
\end{equation}

\begin{equation}
    \begin{aligned}
&\begin{cases}
	if\,\,k_{1}^{*}=k_{2}^{*},S_M\cup k_{2}^{*};\\
	if\,\,k_{1}^{*}\ne k_{2}^{*},\left\{ \begin{array}{c}
	S_I\cup k_{2}^{*},if\,\,\rho _{I}^{k}>\chi\\
	S_M\cup k_{2}^{*},if\,\,\rho _{I}^{k}\leqslant \chi\\
\end{array} \right.\\
\end{cases}
\\
&k_{2}^{*}\in \underset{k\in \mathrm{rand}\left( V\backslash S_M\backslash S_I,\mathrm{s} \right)}{\mathrm{arg}\max}\left[ \bar{G}\left( S_M\cup S_I \right) -\bar{G}\left( \left\{ k \right\} \cup S_M\cup S_I \right) \right] 
    \end{aligned}
    \label{eq:select uni-client}
\end{equation}

For every selection, we randomly sample a subset of clients $s$. A multi-modal clients $k_{1}^{*}$ is selected from $s$ according to $G\left( S_M\right)$ while $k_{2}^{*}$ is also selected from $s$ according to $G\left( S_M\cup S_I \right)$. $k_{1}^{*}=k_{2}^{*}$ means using multi-modal data from this client can contribute most to the global model. When $k_{1}^{*}\ne k_{2}^{*}$, we allocate it to uni-modal or multi-modal client according to its local imbalance ratio: if it is severely imbalanced, we use its uni-weak-modal data for training and aggregation to alleviate the global modal bias, otherwise we believe that training with its multi-modal data contributes more than uni-modal data. $\chi$ is a hyper-parameter.

\noindent\textbf{Discussion.} 
(1) 
Overcoming the modal bias in our method are twofold: the ME loss alleviates imbalance at local side and the selected uni-modal clients further promote balanced learning of global model. Meanwhile, the diversity coming from two submodular functions ensures the representative information for the global model. (2) We assume $I$ is the weak modality above while in practice, we can determine the weak modality before modality selection via the aggregated global imbalance ratio $\rho _I=\frac{1}{\sum\nolimits_{k=1}^N{n_k}}\sum\nolimits_{k=1}^N{\rho _{I}^{k}\cdot n_k}$. Overall, the pseudo-code of BMSFed is provided in \cref{alg:BMSFed}. (3) Only the gradients, prototypes and $\rho_{I}^{k}$ participate in communication, so there is no privacy issue and similar communication overheads as in traditional FL.

\section{Evaluation}
\label{sec:evaluation}
\subsection{Datasets and baselines}
\textbf{Datasets.} We conduct experiments on four datasets: (1) \textbf{CREMA-D} \cite{cao2014crema} is an audio-visual dataset for emotion recognition task with total six categories for emotional states. (2) \textbf{AVE} \cite{tian2018audio} is an audio-visual dataset for event localization with 28 event classes, and here we use it to construct a labeled multi-modal classification dataset following \cite{fan2023pmr}. (3) \textbf{Colored-and-gray MNIST} (CG-MNIST) \cite{kim2019learning} is a synthetic dataset based on MNIST \cite{lecun1998gradient} with gray-scale and monochromatic images as two modalities, following \cite{wu2022characterizing}. (4) \textbf{ModelNet40} is one of the Princeton ModelNet datasets \cite{wu20153d} with 3D objects of 40 categories. The front and back \cite{su2015multi} views are considered as two modalities, following \cite{chen2022fedmsplit}. Details of these four datasets are in the Appendix.

\noindent\textbf{Baselines.}
We choose eight baselines for comparison from four categories: (1) three uni-modal FL methods designed for statistical heterogeneity are extended to multi-modal scenarios: FedAvg \cite{mcmahan2017communication}, FedProx \cite{li2020federatedb} and FedProto \cite{tan2022fedproto}. (2) Integrating OGM-GE \cite{peng2022balanced} and PMR \cite{fan2023pmr}, the solutions for modality imbalance, with FedAvg forms two MFL methods: FedOGM and FedPMR.
(3) Two client selection method: Power-of-choice (pow-d) \cite{cho2020client} and DivFL \cite{balakrishnan2022diverse}, evolved from its uni-modal version directly. (4) One MFL method, FedMSplit \cite{chen2022fedmsplit}, especially designed for modality incongruity.
Compared with these baselines, we demonstrate that an elaborate modality selection strategy is essential to realize comprehensive information exploitation in MFL.

\subsection{Experimental settings}


For CREMA-D, AVE and ModelNet40, we use ResNet18 \cite{he2016deep} as the backbone for audio, visual and flow modalities. Audio data is converted to a spectrogram of size 257x299 for CREMA-D and 257×1,004 for AVE. We randomly choose 3 frames and 4 frames to build image training sets for CREMA-D and AVE respectively.
For CG-MNIST, we build a neural network with 4 convolution layers and 1 average pool layer as the encoder, following the setting as in \cite{fan2023pmr}. We choose the simple yet effective fusion method, concatenation \cite{owens2018audio}, to build fusion classifier for all the datasets.
We set 20 clients for CREMA-D, AVE and ModelNet40 while the number for CG-MNIST is 30. 5 clients are selected in each communication round for CREMA-D, AVE, ModelNet40 and 6 for CG-MNIST. For IID setting, training data is uniformly distributed to all clients. For non-IID scenarios, we use Dirichlet distribution \cite{hsu2019measuring} $Dir\left( \alpha \right) $ to split data ($\alpha=3$ for CREMA-D, AVE, ModelNet40, $\alpha=2$ for CG-MNIST). 
The optimizer is SGD \cite{robbins1951stochastic} for all datasets. Learning rate is initialized at 1e-3 or 1e-2 for CEAMA-D, AVE and ModelNet40 or CG-MNIST and becomes 1e-4 or 1e-3 in the later training stage. The hyper-parameter $\chi$ is set to 1.2-2.5 according to datasets and settings.
To complete stochastic greedy algorithm for \cref{eq:select multi-client,eq:select uni-client}, we use the gradients from the selected clients at current round to update part of the similarity matrix, which is named ``no-overheads'' in \cite{balakrishnan2022diverse}. Except for pow-d, DivFL and BMSFed, other baselines select clients randomly. Each client has two modal data by default.
We do all experiments on a workstation with an RTX 3090 GPU, a 3.9-GHZ Intel Core i9-12900K CPU and 64GB of RAM.

\begin{table}[t]
\renewcommand\arraystretch{1.0}
\caption{Comparison results on four datasets. The metric is the top-1 accuracy (\%). The best is in \textbf{bold}, and the second best is \underline{underlined}. Our method achieves significant improvement on both IID and non-IID scenarios.}
\label{tab:comparison with baselines}
\centering
\setlength{\tabcolsep}{2.mm}{
\begin{tabular}{c|cc|cc|cc|cc}
\hline
Dataset & \multicolumn{2}{c|}{CREMA-D} & \multicolumn{2}{c|}{AVE}  & \multicolumn{2}{c|}{CG-MNIST} & \multicolumn{2}{c}{ModelNet40}    \\ \hline
Method & \multicolumn{1}{c|}{IID}           & non-IID       & \multicolumn{1}{c|}{IID}           & non-IID       & \multicolumn{1}{c|}{IID}           & non-IID       & \multicolumn{1}{c|}{IID} & non-IID \\ \hline
FedAvg    & \multicolumn{1}{c|}{50.7}          & 49.8          & \multicolumn{1}{c|}{62.2}          & 59.7          & \multicolumn{1}{c|}{42.3}          & 41.7          & \multicolumn{1}{c|}{87.2}    &  86.5       \\ 
FedProx   & \multicolumn{1}{c|}{51.0}          & 49.0          & \multicolumn{1}{c|}{62.6}          & 59.9          & \multicolumn{1}{c|}{42.9}          & 43.6          & \multicolumn{1}{c|}{86.9}    &   87.1      \\ 
FedProto  & \multicolumn{1}{c|}{\underline{58.7}}    & 54.0          & \multicolumn{1}{c|}{61.7}          & 58.8          & \multicolumn{1}{c|}{51.5}          & 51.4          & \multicolumn{1}{c|}{87.5}    &  87.2       \\ \hline
FedOGM    & \multicolumn{1}{c|}{56.9}          & \underline{56.4}    & \multicolumn{1}{c|}{62.8}          & 59.3          & \multicolumn{1}{c|}{57.2}          & 53.0          & \multicolumn{1}{c|}{\underline{87.6}}    &  87.0       \\ 
FedPMR    & \multicolumn{1}{c|}{55.5}          & 55.1          & \multicolumn{1}{c|}{63.1}          & \underline{61.6}    & \multicolumn{1}{c|}{\underline{66.1}}    & \underline{63.3}    & \multicolumn{1}{c|}{\underline{87.6}}    &   \textbf{87.7}      \\ \hline
pow-d     & \multicolumn{1}{c|}{50.5}          & 50.7          & \multicolumn{1}{c|}{62.5}          & 60.0          & \multicolumn{1}{c|}{41.2}          & 40.3          & \multicolumn{1}{c|}{86.8}    &  86.2       \\ 
DivFL     & \multicolumn{1}{c|}{51.7}          & 50.8          & \multicolumn{1}{c|}{\underline{63.3}}    & 59.6          & \multicolumn{1}{c|}{43.0}          & 42.1          & \multicolumn{1}{c|}{86.5}    &  86.4       \\ \hline
FedMSplit & \multicolumn{1}{c|}{52.4}          & 51.6          & \multicolumn{1}{c|}{62.4}          & 60.8          & \multicolumn{1}{c|}{43.5}          & 50.9          & \multicolumn{1}{c|}{87.5}    &  87.4       \\ \hline \rowcolor{gray!20}
BMSFed    & \multicolumn{1}{c|}{\textbf{64.5}} & \textbf{61.6} & \multicolumn{1}{c|}{\textbf{64.7}} & \textbf{62.1} & \multicolumn{1}{c|}{\textbf{70.2}} & \textbf{66.7} & \multicolumn{1}{c|}{\textbf{88.7}}    &   \underline{87.5}      \\ \hline
\end{tabular}
}
\end{table}

\begin{figure*}[t]
    \centering
    \begin{subfigure}{0.45\linewidth}
        \includegraphics[width=0.9\linewidth]{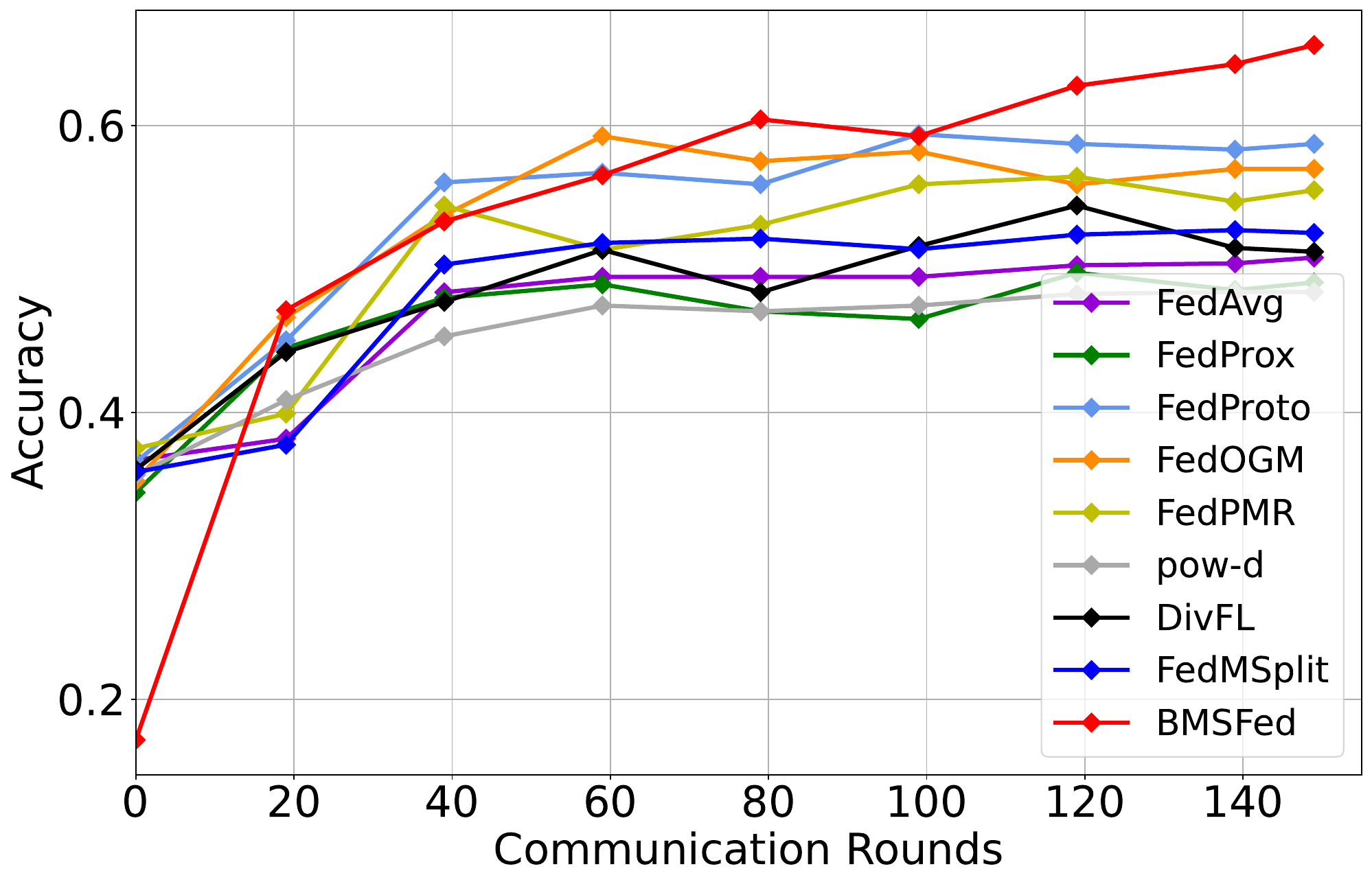}
        \caption{CREMA-D under IID}
        \label{fig:IID CREMAD}
    \end{subfigure}
    \begin{subfigure}{0.45\linewidth}
        \includegraphics[width=0.9\linewidth]{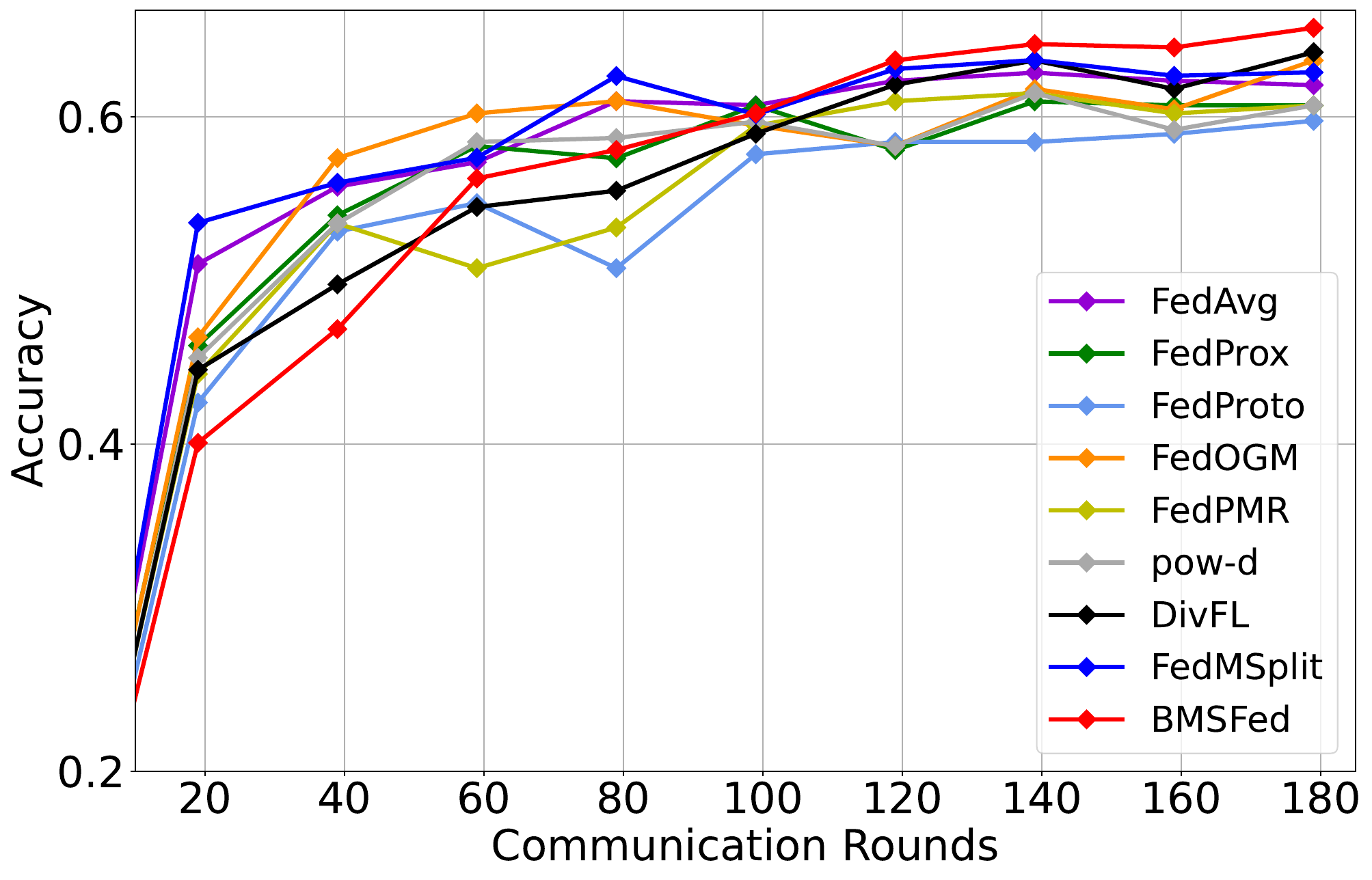}
        \caption{AVE under IID}
        \label{fig:IID AVE}
    \end{subfigure}
    
    \begin{subfigure}{0.45\linewidth}
        \includegraphics[width=0.9\linewidth]{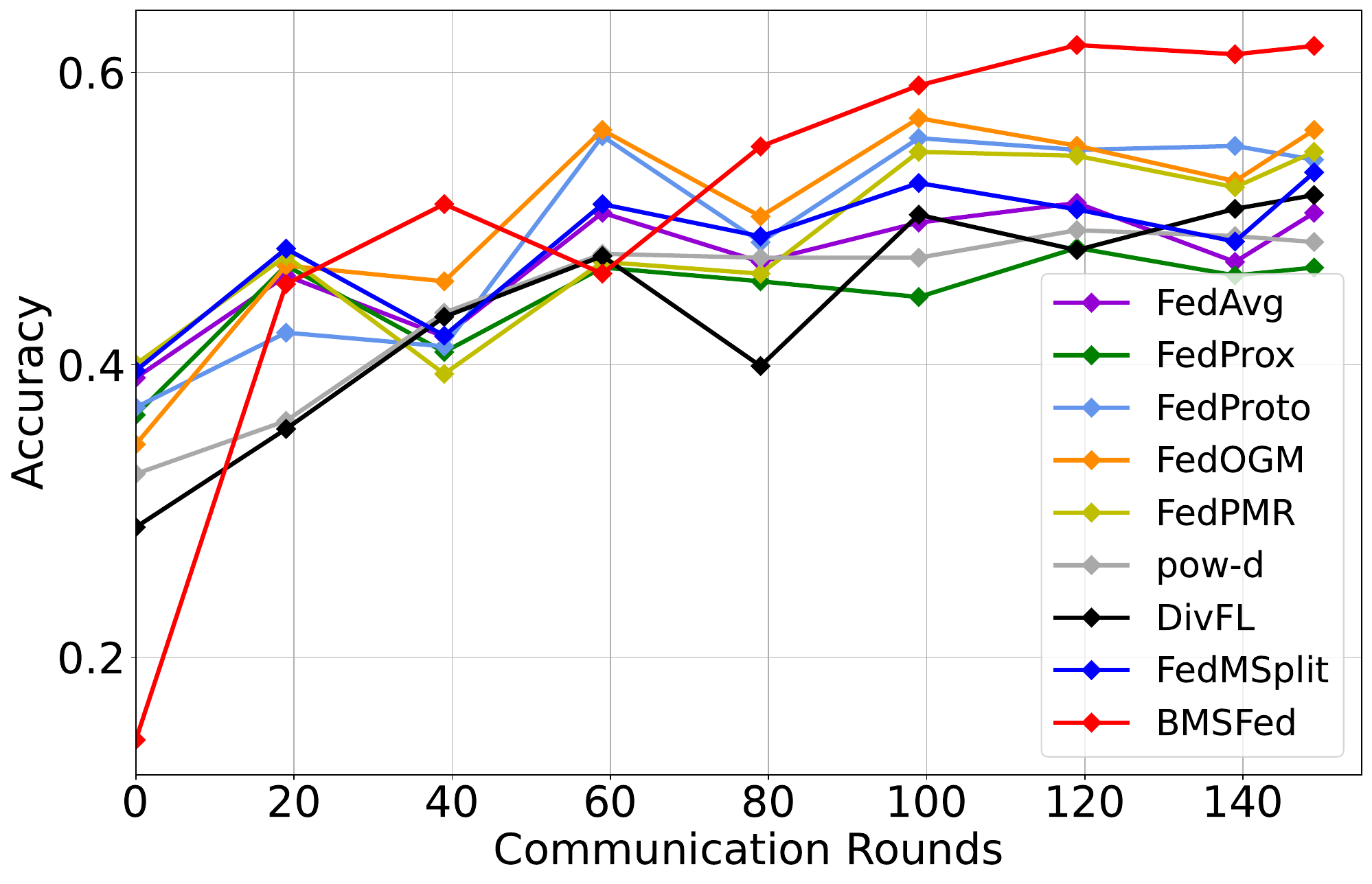}
        \caption{CREMA-D under non-IID}
        \label{fig:non-IID CREMAD}
    \end{subfigure}
    \begin{subfigure}{0.45\linewidth}
        \includegraphics[width=0.9\linewidth]{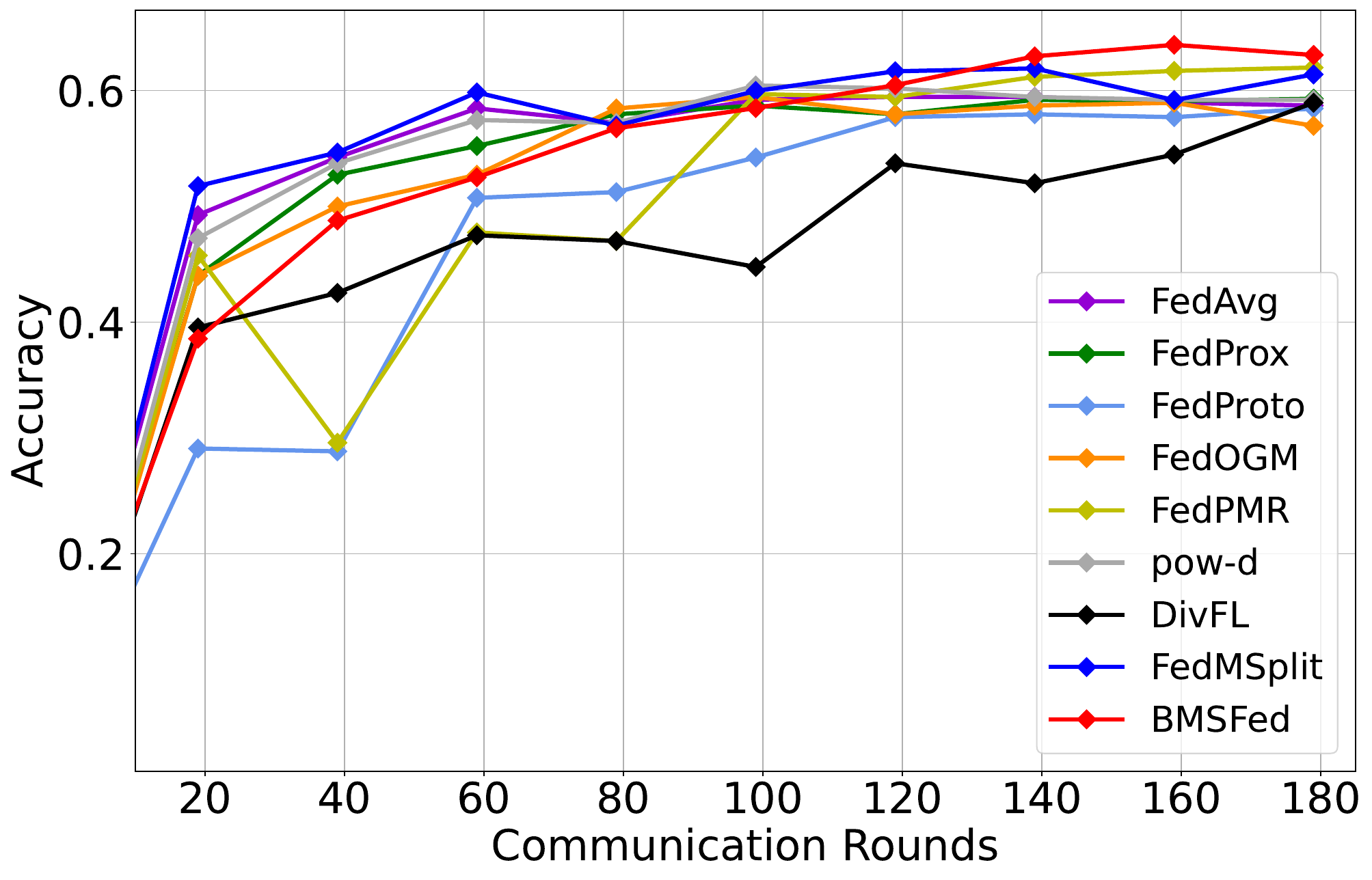}
        \caption{AVE under non-IID}
        \label{fig:non-IID AVE}
    \end{subfigure}
    \caption{Test accuracy of BMSFed compared with other baselines on CREMA-D and AVE. BMSFed converges to more accurate solutions than all baselines.}
    \label{fig:training curve}
\end{figure*}

\begin{table}[t]
\renewcommand\arraystretch{1.0}
\caption{The uni-modal performance comparison on CREMA-D and AVE. The metric is the top-1 accuracy (\%). Our method achieves the best uni-visual (the weak modality) performance on all settings and maintains the comparable uni-audio performance.}
\label{tab:unimodal comparison}
\centering
\setlength{\tabcolsep}{2.2mm}{
\begin{tabular}{c|cccc|cccc}
\hline
Dataset   & \multicolumn{4}{c|}{CREMA-D}                                                                                                 & \multicolumn{4}{c}{AVE}                                                                                                      \\ \hline
Setting   & \multicolumn{2}{c|}{IID}                                                & \multicolumn{2}{c|}{non-IID}                       & \multicolumn{2}{c|}{IID}                                                & \multicolumn{2}{c}{non-IID}                        \\ \hline
Method    & \multicolumn{1}{c|}{A}             & \multicolumn{1}{c|}{V}             & \multicolumn{1}{c|}{A}             & V             & \multicolumn{1}{c|}{A}             & \multicolumn{1}{c|}{V}             & \multicolumn{1}{c|}{A}             & V             \\ \hline
FedAvg    & \multicolumn{1}{c|}{51.2}          & \multicolumn{1}{c|}{20.6}          & \multicolumn{1}{c|}{50.7}          & 20.2          & \multicolumn{1}{c|}{61.1}          & \multicolumn{1}{c|}{26.8}          & \multicolumn{1}{c|}{61.4}          & 26.4          \\
FedProx   & \multicolumn{1}{c|}{51.3}          & \multicolumn{1}{c|}{20.2}          & \multicolumn{1}{c|}{50.1}          & 22.0          & \multicolumn{1}{c|}{60.4}          & \multicolumn{1}{c|}{27.1}          & \multicolumn{1}{c|}{61.2}          & 26.9          \\
FedProto  & \multicolumn{1}{c|}{50.2}          & \multicolumn{1}{c|}{35.3}          & \multicolumn{1}{c|}{48.6}          & \underline{39.1}    & \multicolumn{1}{c|}{55.7}          & \multicolumn{1}{c|}{36.8}          & \multicolumn{1}{c|}{59.7}          & 32.8          \\ \hline
FedOGM    & \multicolumn{1}{c|}{50.5}          & \multicolumn{1}{c|}{35.7}          & \multicolumn{1}{c|}{48.8}          & 30.2          & \multicolumn{1}{c|}{58.7}          & \multicolumn{1}{c|}{28.8}          & \multicolumn{1}{c|}{59.4}          & 29.4          \\
FedPMR    & \multicolumn{1}{c|}{51.5}          & \multicolumn{1}{c|}{\underline{38.7}}    & \multicolumn{1}{c|}{50.1}          & 35.9          & \multicolumn{1}{c|}{61.7}          & \multicolumn{1}{c|}{\underline{39.6}}    & \multicolumn{1}{c|}{\underline{61.7}}    & \underline{35.3}    \\ \hline
pow-d     & \multicolumn{1}{c|}{51.5}          & \multicolumn{1}{c|}{20.4}          & \multicolumn{1}{c|}{\underline{51.6}}    & 18.8          & \multicolumn{1}{c|}{\underline{61.9}}    & \multicolumn{1}{c|}{26.9}          & \multicolumn{1}{c|}{60.1}          & 27.1          \\
DivFL     & \multicolumn{1}{c|}{\textbf{52.3}} & \multicolumn{1}{c|}{21.1}          & \multicolumn{1}{c|}{\textbf{52.1}} & 22.7          & \multicolumn{1}{c|}{\textbf{62.7}} & \multicolumn{1}{c|}{25.3}          & \multicolumn{1}{c|}{61.6}          & 26.3          \\ \hline
FedMSplit & \multicolumn{1}{c|}{\underline{52.0}}    & \multicolumn{1}{c|}{21.8}          & \multicolumn{1}{c|}{50.8}          & 21.6          & \multicolumn{1}{c|}{61.3}          & \multicolumn{1}{c|}{26.9}          & \multicolumn{1}{c|}{\textbf{62.3}} & 28.7          \\ \hline \rowcolor{gray!20}
BMSFed    & \multicolumn{1}{c|}{51.0}          & \multicolumn{1}{c|}{\textbf{41.9}} & \multicolumn{1}{c|}{49.3}          & \textbf{41.4} & \multicolumn{1}{c|}{59.7}          & \multicolumn{1}{c|}{\textbf{40.2}} & \multicolumn{1}{c|}{60.2}          & \textbf{38.6} \\ \hline
\end{tabular}
}
\end{table}

\subsection{Comparison with baselines}
\noindent\textbf{BMSFed effectively improves the performance.} As demonstrated in \cref{tab:comparison with baselines}, our BMSFed achieves the best results on the four datasets under both IID and non-IID settings (by up to 5.8\% on CREMA-D). Client sampling here (pow-d and DivFL) cannot fully exploit information for all modalities, making its improvement limited or even worse than FedAvg in CG-MNIST. Although FedOGM and FedPMR accomplish modest improvement because of their ability to alleviate modality imbalance, they are not as good as BMSFed since they do not consider the overall performance of the global model. Traditional uni-modal FL methods for statistical heterogeneity (e.g. FedProx) and MFL method for modality incongruity (FedMSplit) only obtain slight improvement. We also illustrate the trend of test accuracy versus the number of communication rounds on CREMA-D and AVE in \cref{fig:training curve}. BMSFed can realize comparable or even faster convergence speeds in CREMA-D and AVE.

\noindent\textbf{BMSFed exploits all modalities comprehensively.} To show the effect of our method on addressing modal bias, 
we report the performance of each modality on CREMA-D and AVE as shown in \cref{tab:unimodal comparison}. The uni-modal performance evaluation follows \cite{fan2023pmr}: a sample is classified into the class corresponding to its nearest prototype. It is clear that BMSFed could considerably improve the performance of weak modality (visual) and mitigate the modality-level bias. Besides, compared with randomly modality abandoning, which significantly reduces audio performance as illustrated in \cref{tab:FL modal bias}, BMSFed achieves comparable audio performance with other baselines. 
Although FedProto, FedOGM and FedPMR also alleviate the imbalance, they mainly focus on local optimization, resulting in the performance gap between them and our BMSFed on the aggregated model, which further indicates that in MFL, it is important to take both local optimization for each modality and the overall performance for global model into consideration simultaneously.

\begin{table}[t]
\renewcommand\arraystretch{1.0}
\caption{Ablation study. `BMSFed-local' uses local prototypes rather than global prototypes for ME loss.}
  \label{tab:ablation}
  \centering
  \setlength{\tabcolsep}{3.5mm}{
  \begin{tabular}{ c | c c | c c  }
    \hline
    Dataset & \multicolumn{2}{c|}{CREMA-D} & \multicolumn{2}{c}{AVE} \\
    \hline
    setting & IID & non-IID & IID & non-IID \\
    \hline
    FedAvg & 50.7 & 49.8 & 62.2 & 59.7 \\
    DivFL & 51.7 & 50.8 & 63.3 & 59.6 \\
    \hline
    FedAvg-0.2 & 52.4 & 50.1 & 63.4 & 61.1 \\
    FedAvg-0.5 & 55.7 & 55.1 & 60.7 & 59.4 \\
    FedAvg-0.8 & 61.2 & 58.1 & 58.5 & 58.7 \\
    \hline
    FedAvg+ME & 55.8 & 54.5 & 62.8 & 60.7 \\
    DivFL+ME & 57.1 & 55.6 & 63.0 & 61.1 \\
    BMSFed-local & 63.7 & 60.3 & 63.4 & 60.1 \\
    \hline \rowcolor{gray!20}
    BMSFed & \textbf{64.5} & \textbf{61.6} & \textbf{64.7} & \textbf{62.1} \\
    \hline
  \end{tabular}
  }
\end{table}

\subsection{Ablation study}
\noindent\textbf{Effectiveness of each component.} \cref{tab:ablation} studies the effect of each BMSFed component. Applying ME loss \cref{eq:global PCE} on random sampling (FedAvg+ME) and the well-designed client selection (DivFL+ME) surpasses their vanilla strategies (FedAvg, DivFL) by a large margin, demonstrating its effectiveness on local enhancement. Comparing BMSFed (64.5\% on IID CREMA-D) with `DivFL+ME' (57.1\% on the same setting) also denotes the necessity of balancing different modalities considering the global model via modality selection. To show the importance of aligning feature spaces of weak modality, we replace the global prototypes with local prototypes (BMSFed-local). Global alignment achieves notable improvement (by up to 2\% on non-IID AVE). The performance improvement compared with `FedAvg-0.2,0.5,0.8' 
exhibits that randomly sampling modalities does not always lead to improvement
and further demonstrates the need of meticulously selecting modalities for information exploitation.

\begin{figure*}[t]
    \centering
    \begin{subfigure}{0.325\linewidth}
        \includegraphics[width=0.9\linewidth]{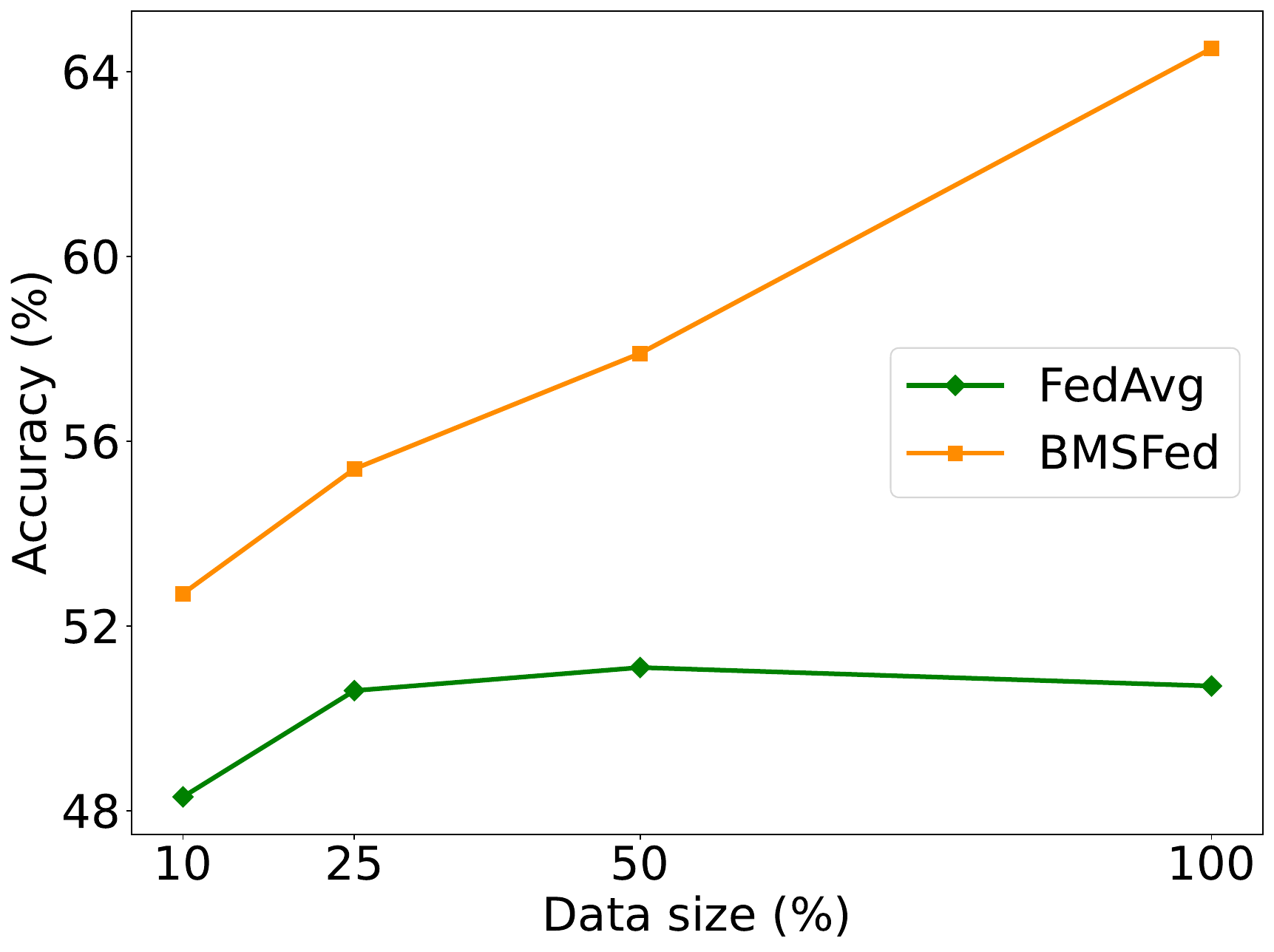}
        \caption{CREMA-D under IID}
        \label{fig:data size}
    \end{subfigure}
    \begin{subfigure}{0.325\linewidth}
        \includegraphics[width=0.9\linewidth]{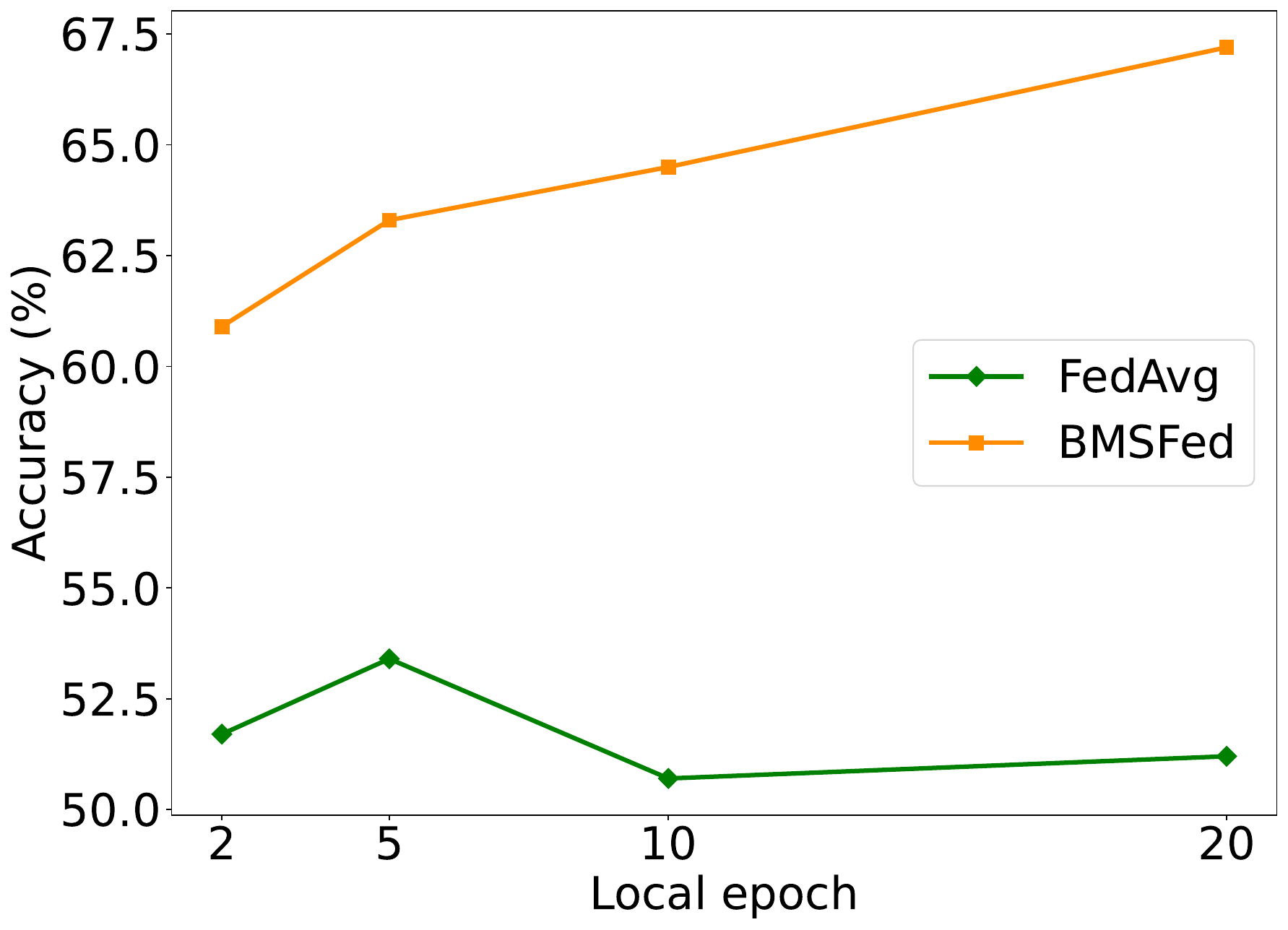}
        \caption{CREMA-D under IID}
        \label{fig:local epoch}
    \end{subfigure}
    \begin{subfigure}{0.325\linewidth}
        \includegraphics[width=0.9\linewidth]{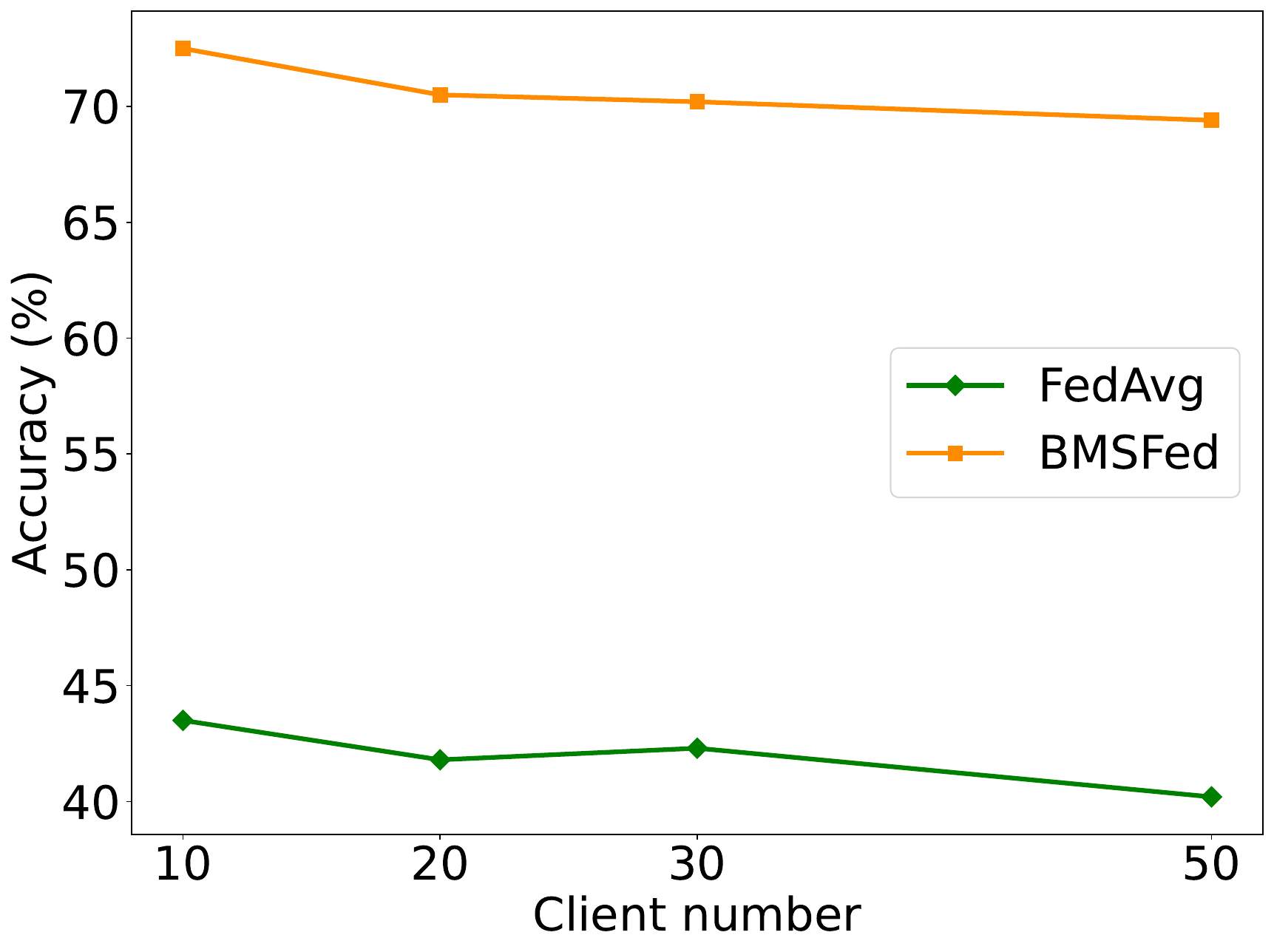}
        \caption{CG-MNIST under IID}
        \label{fig:client number}
    \end{subfigure}
    \caption{Robustness validation on data size, local epoch and client number. Our BMSFed consistently outperforms baseline (FedAvg) under various scenarios.}
    \label{fig:robustness}
\end{figure*}

\begin{figure}[t]
    \centering
    \includegraphics[width=0.5\linewidth]{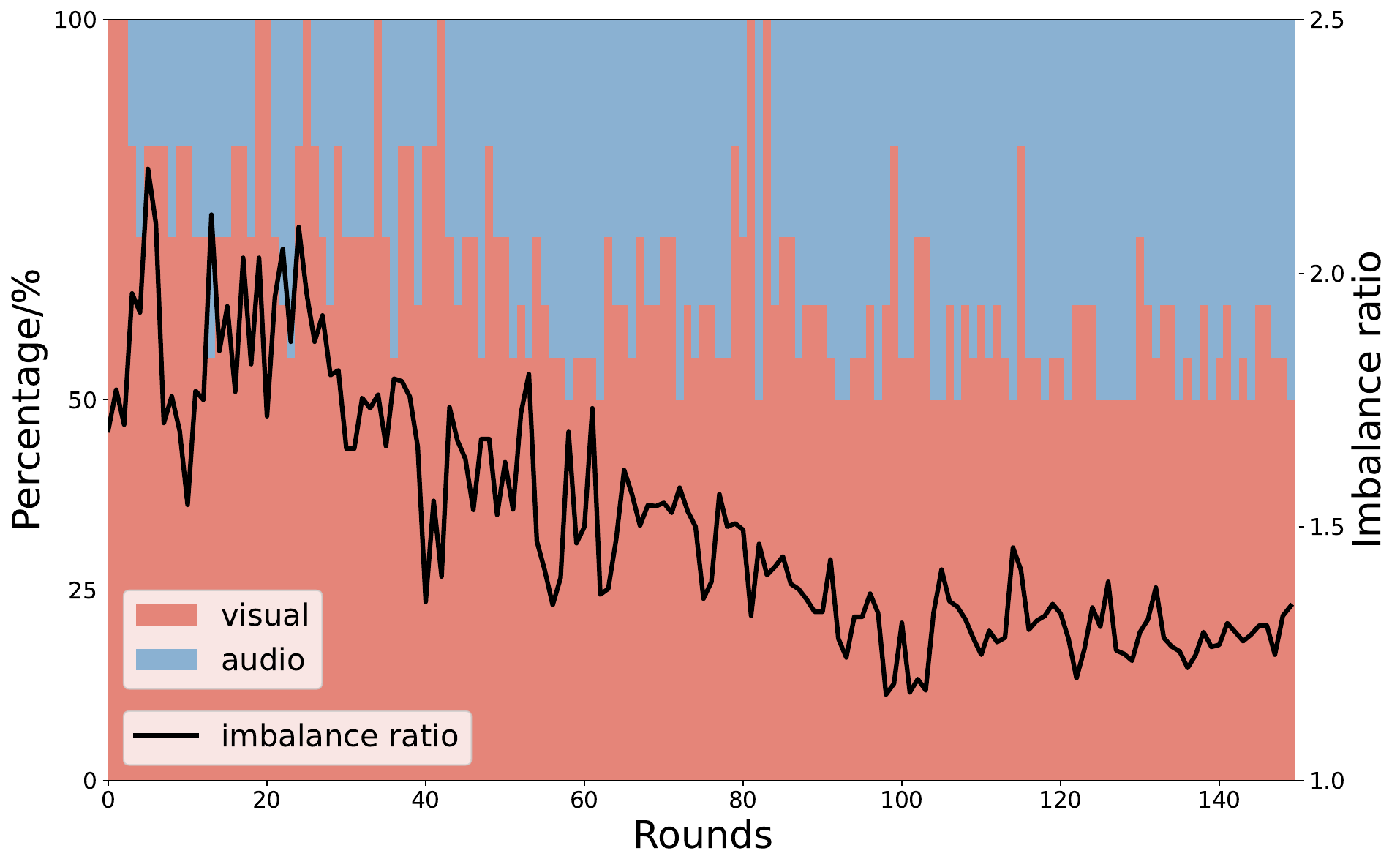}    
    \caption{Proportional change of audio and visual respectively and the curve of global imbalance ratio during training on CREMA-D under IID setting.}
    \vspace{-14pt}
    \label{fig:ratio}
\end{figure}

\noindent\textbf{Robustness test.} To verify the robustness of our method, we vary three key hyperparameters to build various scenarios: (1) change the data size $\left| \mathcal{D} _i \right|$ to allow each client to hold a small amount of data, (2) set different local training epochs, and (3) vary the total client number $N$. As illustrated in \cref{fig:data size,fig:local epoch}, our BMSFed consistently outperforms baseline (FedAvg) under various scenarios. More data as well as more local training epochs can bring further improvements to our method, indicating that exploiting the weak modality need more training efforts. Based on the results in \cref{fig:client number}, our approach can also be generalised to clients with larger scales.

\noindent\textbf{The relationship between modality selection and imbalance degree.} In \cref{eq:select multi-client,eq:select uni-client}, we use a conflict resolution strategy based on local imbalance ratio to realize balanced modality selection. We visualize the proportions of audio and visual modalities selected in each round 
and the global imbalance ratio. It is clear from \cref{fig:ratio} that audio is the dominant modality (imbalance ratio is always greater than 1) and modality imbalance is gradually alleviated as training progresses (global imbalance ratio shows a downward trend). In addition, the proportion of selected visual modality follows the same trend (larger in the early stage of training and becomes smaller later), implying the rationality of our selection strategy based on imbalance ratio and its effectiveness on mitigating bias.

\noindent\textbf{Effectiveness on modality incongruity scenario.} All above experiments assume that each client initially has complete modal data.
Here, we build the stimulation of modality incongruity scenario, in which half of the clients have data with two modalities, and the other half only have data in one modality: random audio or visual. The results are shown in \cref{tab:drop setting} 
(The results of pow-d is not available because it is not applicable to this scenario).
Our BMSFed still makes impressive improvement compared with all other baselines (by up to 3.5\% on IID CREMA-D), illustrating the good generalization ability of our method in different scenarios. It is worth mentioning that FedMSplit performs better than before because it is specifically designed for modality incongruity.

\begin{table}[t]
\renewcommand\arraystretch{1.1}
\caption{Performance on CREMA-D and AVE with modality incongruity. 50\% of clients have all modal data and 50\% of clients only retain data with a single modality (audio or visual).}
  \label{tab:drop setting}
  \centering
  \setlength{\tabcolsep}{3.5mm}{
  \begin{tabular}{ c | c c | c c }
    \hline
    Dataset & \multicolumn{2}{c|}{CREMA-D} & \multicolumn{2}{c}{AVE} \\
    \hline
    setting & IID & non-IID & IID & non-IID \\
    \hline
    FedAvg & 55.7 & 55.1 & 60.7 & 58.7 \\
    \hline
    FedProx & 56.8 & 56.0 & 61.2 & 58.5 \\
    FedProto & 58.7 & 57.0 & 61.3 & 59.7 \\
    FedOGM & 58.6 & 57.4 & 60.1 & 58.5 \\
    FedPMR & 56.4 & 55.5 & 61.9 & 60.3 \\
    DivFL & 57.4 & 55.6 & 61.1 & 58.6 \\
    FedMSplit & 58.9 & 56.9 & 61.7 & 60.0 \\
    \hline \rowcolor{gray!20}
    BMSFed & \textbf{62.4} & \textbf{59.8} & \textbf{63.5} & \textbf{60.9} \\
    \hline 
  \end{tabular}
  }
\end{table}

\section{Conclusion}
In this paper, we analyze traditional client selections and find their ineffectiveness in MFL. We further reveal that there exists strong modality-level bias due to the modality imbalance during the training iterations and uni-modal training on some clients may contribute more to the global model than multi-modal training.
To address this issue, we propose the balanced modality selection scheme for MFL (BMSFed) with modality-level gradient decoupling to release the potential of all modalities and maximize the gradient diversities to improve global aggregation.
We also introduce a modal enhancement loss to optimize the local update process.
Our method does not introduce additional local training costs and communication overheads compared with previous methods.
Extensive experiments on four datasets demonstrate the superiority of our method in performance and applicability under different modal combinations, data distributions and modality incongruity scenarios. 


\section*{Acknowledgements}
Our work was supported by two grant from the Research Grants Council of the Hong Kong Special Administrative Region, China (Project No. PolyU15222621, PolyU15225023) and the funding from National Natural Science Foundation of China under grants 62302184.


%
%
\bibliographystyle{splncs04}
\bibliography{main}

\clearpage

\section{Appendix}
\subsection{Datasets}
\textbf{CREMA-D} \cite{cao2014crema} is an audio-visual dataset for emotion recognition task, each video in which consists of both facial and acoustic emotional expressions. There are total 6 categories for emotional states: \textit{neutral}, \textit{happy}, \textit{sad}, \textit{fear}, \textit{disgust} and \textit{anger}. 7,442 clips in total are collected in this dataset. 6,698 samples are randomly chosen as the training set and the rest of 744 samples are the testing set.

\noindent\textbf{AVE} \cite{tian2018audio} is an audio-visual dataset for audio-visual event localization, which consists of 28 event classes and 4,143 10-second video clips with both auditory and visual tracks as well as second-level annotations. All the video clips are collected from YouTube. In our experiments, we aim to construct a labeled multi-modal classification dataset by extracting the frames from event-localized video segments and capturing the audio clips within the same segments. The training and validation splits of the dataset follow \cite{tian2018audio}.

\noindent\textbf{Colored-and-gray MNIST} \cite{kim2019learning} is a synthetic dataset based on MNIST \cite{lecun1998gradient}, and we denote it as CG-MNIST in this paper. Two kinds of images form a sample pair: a gray-scale image and a monochromatic image. In the training set, there are 60,000 sample pairs, and the monochromatic images are strongly color-correlated with their digit labels, In the validation set, the number of sample pairs is 10,000, while the monochromatic images are weakly color-correlated with their labels. The data synthesis method follows \cite{wu2022characterizing}. This dataset is used to prove the method's effectiveness beyond audio-visual modality.

\noindent\textbf{ModelNet40} is one of the Princeton ModelNet datasets \cite{wu20153d} with 3D objects of 40 categories (9,843 training samples and 2,468 test samples). The task is to classify a 3D object based on the 2D views of its front and back \cite{su2015multi}. Each example is a collection of 2D images (224×224 pixels) of a 3D object. 

\subsection{Formula details}
\label{sec:formula details}
Here, we give the calculation details for some formulas which are not discussed in detail in the main paper.

\noindent\textbf{Prototype.} The prototype is the centroid of the representations for each class. Therefore, the local prototype for class $j$ is calculated as (for modal $I$):
\begin{equation}
    c_{j}^{I}=\frac{1}{N_j}\sum\nolimits_{i=1}^{N_j}{z_{j_i}^{I}}
    \label{eq:local proto cal}
\end{equation}
where $N_j$ is the number of samples with class $j$ in local side.

When the sever receives the local prototypes from each client, the global prototypes are aggregated according to the sample numbers from each client:
\begin{equation}
    c_{j}^{GI}=\frac{1}{\sum\nolimits_{k=1}^N{N_{j}^{}|_k}}\sum\nolimits_{k=1}^N{c_{j}^{I}|_k\cdot N_{j}^{}|_k}
    \label{eq:global proto cal}
\end{equation}
where $N_{j}|_k$ is the number of samples with class $j$ in client $k$, $c_{j}^{I}|_k$ is the prototype of class $j$ in client $k$.

\noindent\textbf{Imbalance ratio $\rho_I$ and coef $\beta^k$.} In this paper, we need the coefficients $\beta^k$ and $\gamma^k$ in \cref{eq:local loss} to adjust the strength of local enhancement and imbalance ratio to determine which modality is weak and modulate the modality selection process as in \cref{eq:select uni-client}. Local imbalance ratio of client $k$ is the quotient of average ground-truth logits from two modalities:
\begin{equation}
\begin{aligned}
    \begin{array}{c}
	s_{i}^{A}=\sum_{c=1}^C{1_{c=y_i}\cdot \mathrm{soft}\max \left( \hat{y}_{i}^{A} \right) _c}\\
	s_{i}^{I}=\sum_{c=1}^C{1_{c=y_i}\cdot \mathrm{soft}\max \left( \hat{y}_{i}^{I} \right) _c}\\
\end{array}
\end{aligned}
    \label{eq:logit cal}
\end{equation}
\begin{equation}
    \rho _{I}^{k}=\frac{\sum\nolimits_{i\in B_t}^{}{s_{i}^{A}}}{\sum\nolimits_{i\in B_t}^{}{s_{i}^{I}}}
    \label{eq:local ratio cal}
\end{equation}
where $\hat{y}_{i}^{A},\hat{y}_{i}^{I}$ are the logit outputs based on the distance differences from local prototypes. $B_t$ is a random mini-batch at time step $t$. Then the global imbalance ratio is calculated as:
\begin{equation}
    \rho _I=\frac{1}{\sum\nolimits_{k=1}^N{n_k}}\sum\nolimits_{k=1}^N{\rho _{I}^{k}\cdot n_k}
    \label{eq:global ratio cal}
\end{equation}

According to \cite{fan2023pmr}, we design $\beta^k$ as:
\begin{equation}
    \left\{ \begin{matrix}
	\gamma^k =clip\left( 0,\frac{1}{\rho _{I}^{k}}-1,1 \right) &		\rho _{I}^{k}<1\\
	\beta^k =clip\left( 0,\rho _{I}^{k}-1,1 \right) &		\rho _{I}^{k}\geqslant 1\\
\end{matrix} \right. 
\label{eq:beta cal}
\end{equation}

\subsection{Solving submodular function}
Here is the procedure for solving submodular function with stochastic greedy algorithm. \\
\noindent(1) \textbf{If} global epoch == 1: \\
Broadcast the global model to all clients and update one step (or one epoch) via local loss \cref{eq:local loss} (with local prototypes here) to get local gradients, prototypes and imbalance ratio, \\
            \textbf{else}: \\
             Broadcast the global model to selected clients and update several epochs via local loss \cref{eq:local loss} (with global prototypes here) to get local gradients, prototypes and imbalance ratio; \\
\noindent(2) Gather the local information from all clients and then construct two gradient similarity matrices with the size $N\times N$ for both multi-modal gradients and uni-modal enhancing gradients as shown in \cref{eq:modality selection decoupling}, and also global prototypes and global ratio;  \\
\noindent(3) Randomly select a subset of clients with size $s$ from $V\backslash S_M\backslash S_I$; \\
\noindent(4) Determine the clients with the smallest distance summation from other unselected clients for the two similarity matrices; \\
\noindent(5) Select modalities according to \cref{eq:select multi-client,eq:select uni-client}; \\
\noindent(6) Update the gradient similarity matrices; \\
\noindent(7) Loop (3)-(6) until the number of selected clients reaches upper limit and the selected modalities are finally decided. \\



\subsection{Experiment setting}
\noindent\textbf{Device.} All of our experiments were performed on a NVIDIA GeForce RTX 3090 GPU. We set the same seed for both devices to ensure they obtain the consistent results. The batch size for CREMA-D, AVE and ModelNet40 is 64 and 128 for CG-MNIST.

\noindent\textbf{Data split.} We use Dirichlet distribution to split data on clients. Here we illustrate the visualization of class numbers on CREMA-D. Not only are the total number of samples inconsistent between clients, but the numbers of samples in each class are also different.

\begin{figure}[h]
    \centering
    \begin{subfigure}{0.47\linewidth}
        \includegraphics[width=1.0\linewidth]{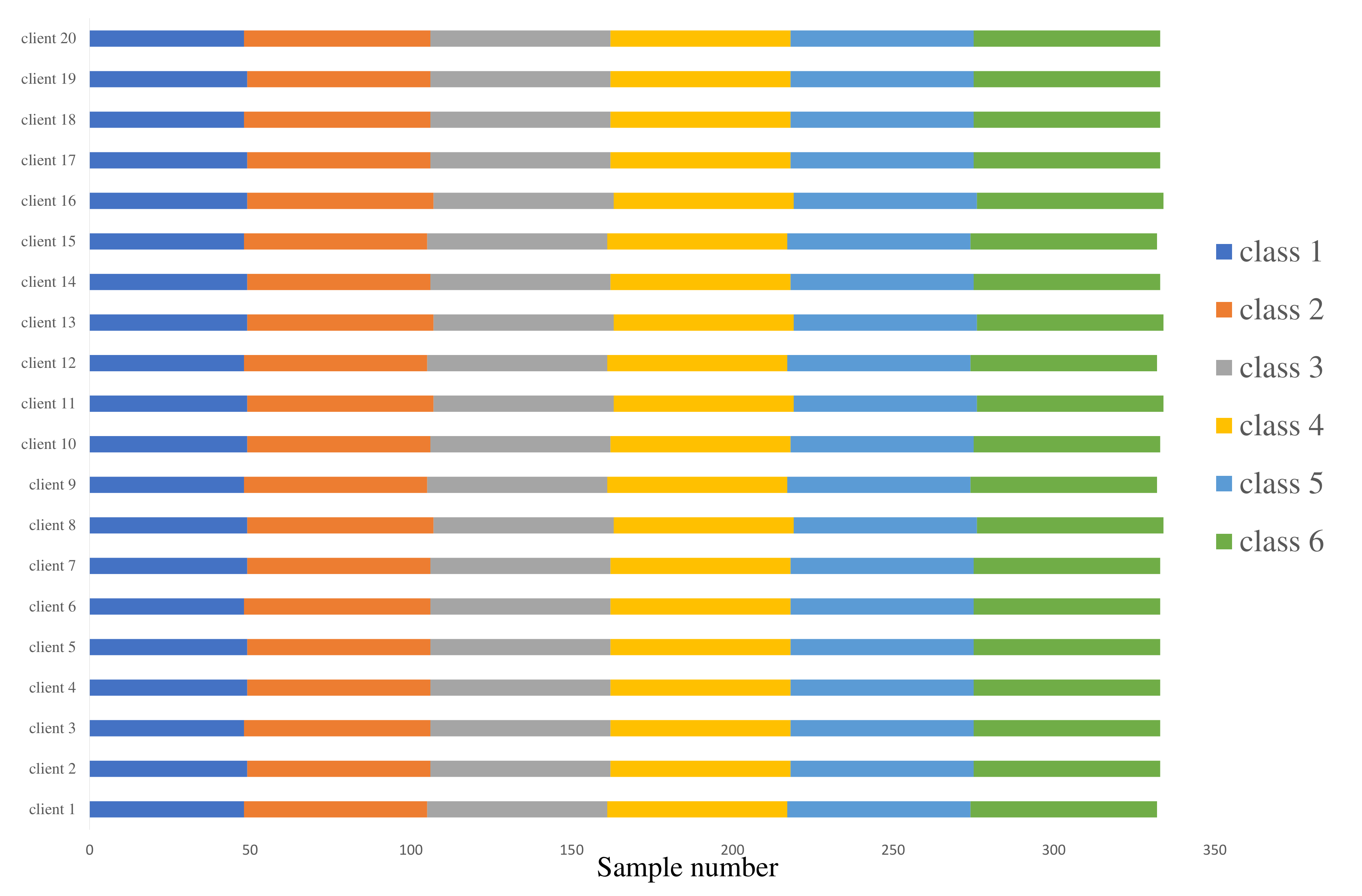}
        \caption{IID}
        \label{fig:audio acc CREMAD}
    \end{subfigure}
    \hfill
    \begin{subfigure}{0.47\linewidth}
        \includegraphics[width=1.0\linewidth]{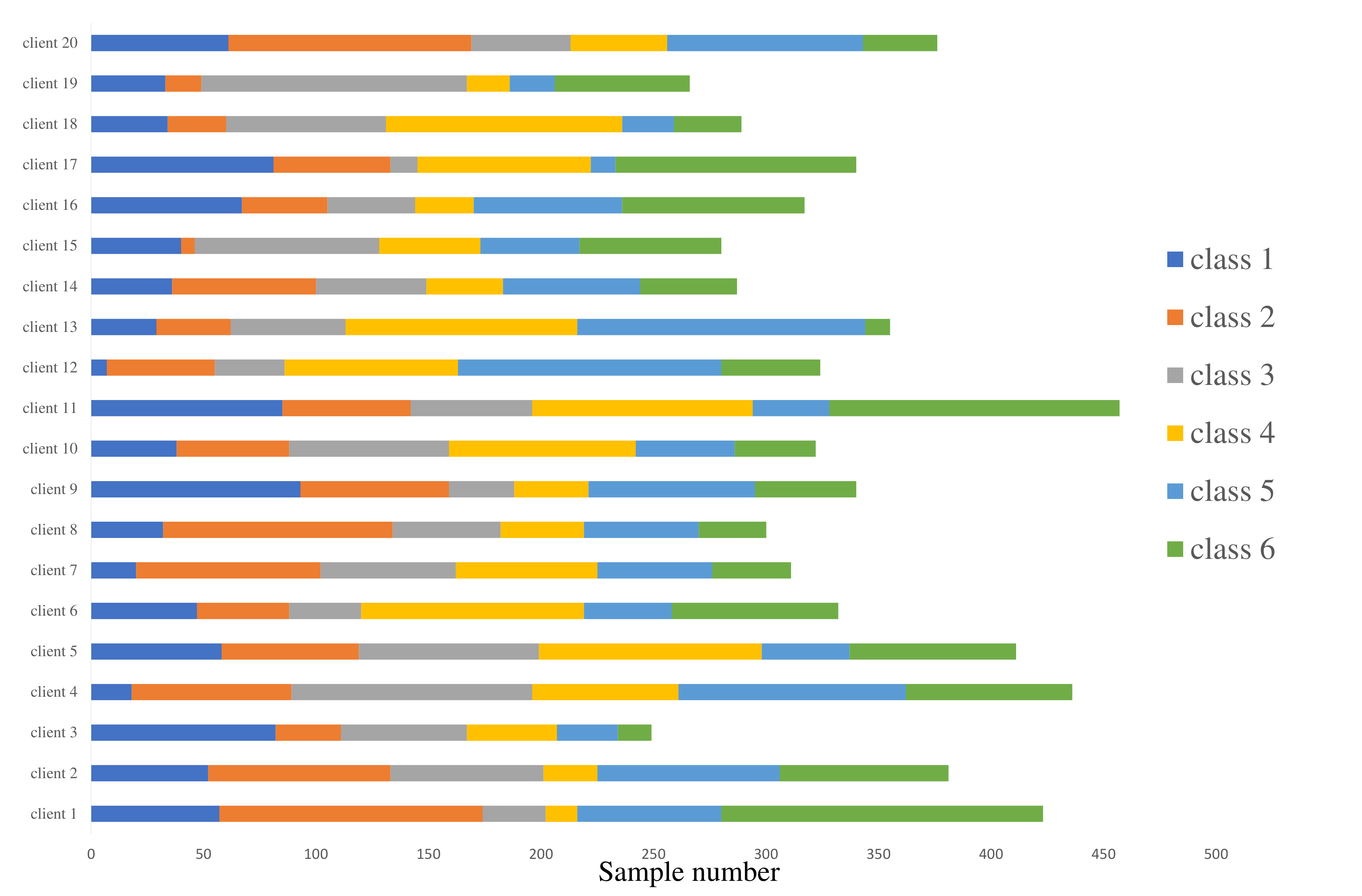}
        \caption{Non-IID}
        \label{fig:visual acc CREMAD}
    \end{subfigure}
    \caption{The visualization of data distribution on CREMA-D.}
    \label{fig:modality performance}
\end{figure}

\noindent\textbf{Implementation of baselines.} \\
In this paper, we use eight baselines for comparison and we describe their implementation here.

For FedAvg, FedProx and FedProto, they are extended to multi-modal FL directly: perform local update based on their algorithms according to the local data and aggregation based on the sample numbers from clients. The hyper-parameter $\mu$ for FedProx is 1.0-2.0 for the three datasets according to different settings. The global prototypes for two modalities are only used in FedProto.

For FedOGM and FedPMR,  we integrate FedAvg and OGM-GE and PMR respectively. They modulate the learning paces of two modalities according to local information (without any global information).

For pow-d, we randomly choose half of all clients (10 for CREMA-D, AVE and ModelNet40, 15 for CG-MNIST) to calculate their losses and select clients from them. For DivFL, it can be extended to multi-modal scenario straightforwardly.

For FedMSplit, $\gamma$ is set to 0.9 for all datasets.

\subsection{Other experimental results}
\begin{figure}[h]
    \centering
    \begin{subfigure}{0.45\linewidth}
        \includegraphics[width=0.9\linewidth]{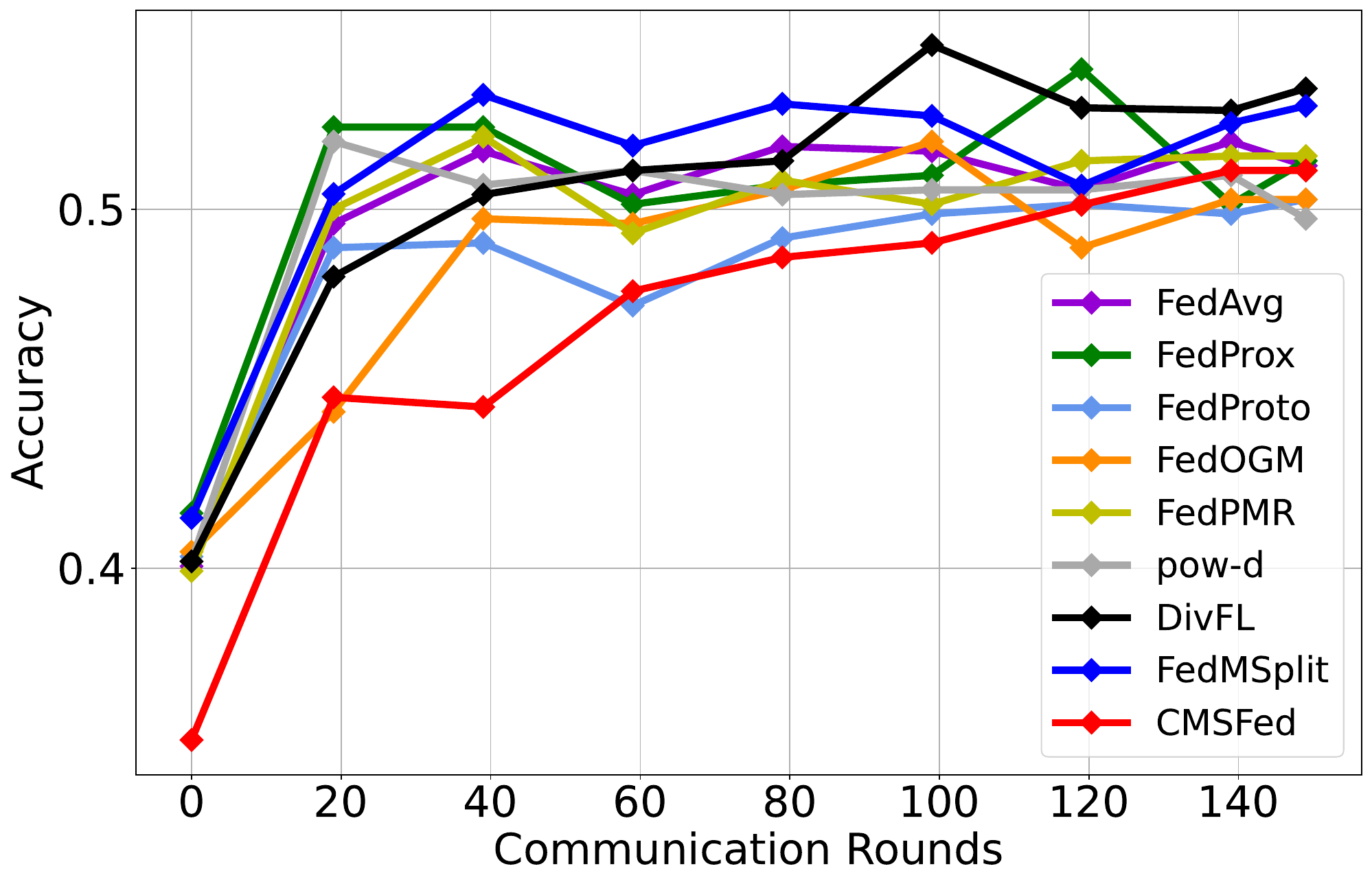}
        \caption{Audio in non-IID CREMA-D}
        \label{fig:audio acc CREMAD non-iid}
    \end{subfigure}
    \begin{subfigure}{0.45\linewidth}
        \includegraphics[width=0.9\linewidth]{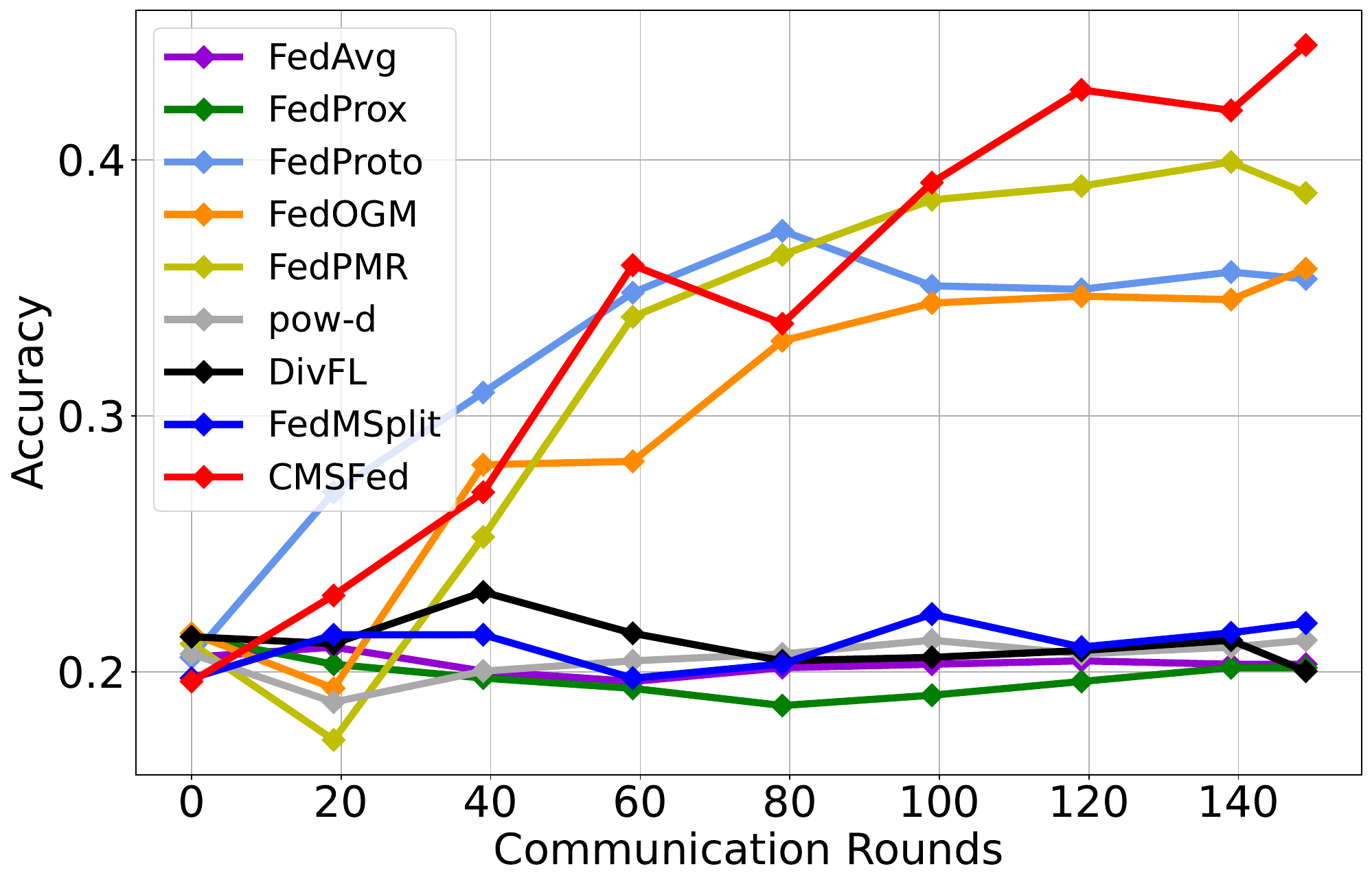}
        \caption{Visual in non-IID CREMA-D}
        \label{fig:visual acc CREMAD non-iid}
    \end{subfigure}
    
    \begin{subfigure}{0.45\linewidth}
        \includegraphics[width=0.9\linewidth]{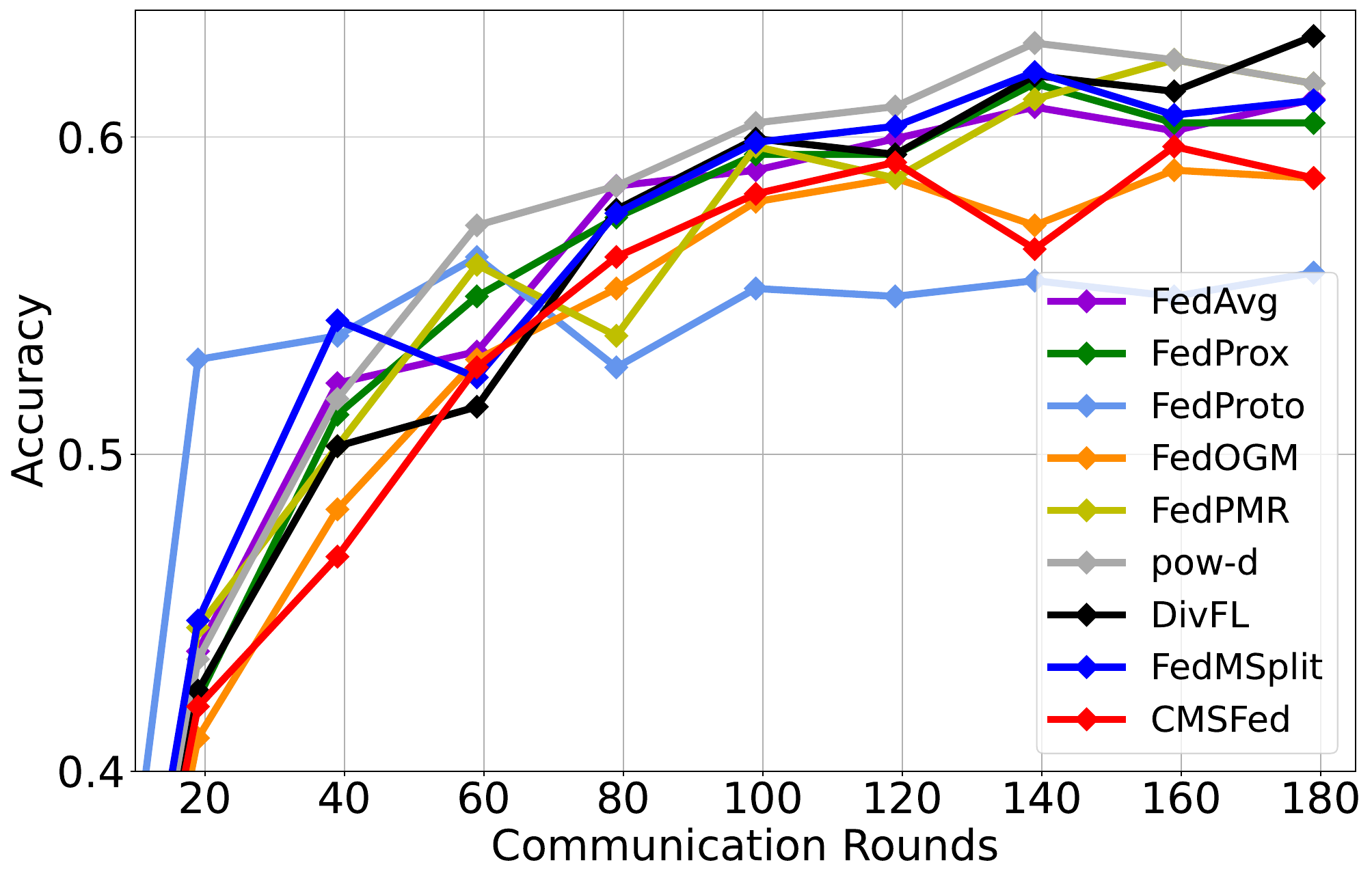}
        \caption{Audio in non-IID AVE}
        \label{fig:audio acc AVE}
    \end{subfigure}
    \begin{subfigure}{0.45\linewidth}
        \includegraphics[width=0.9\linewidth]{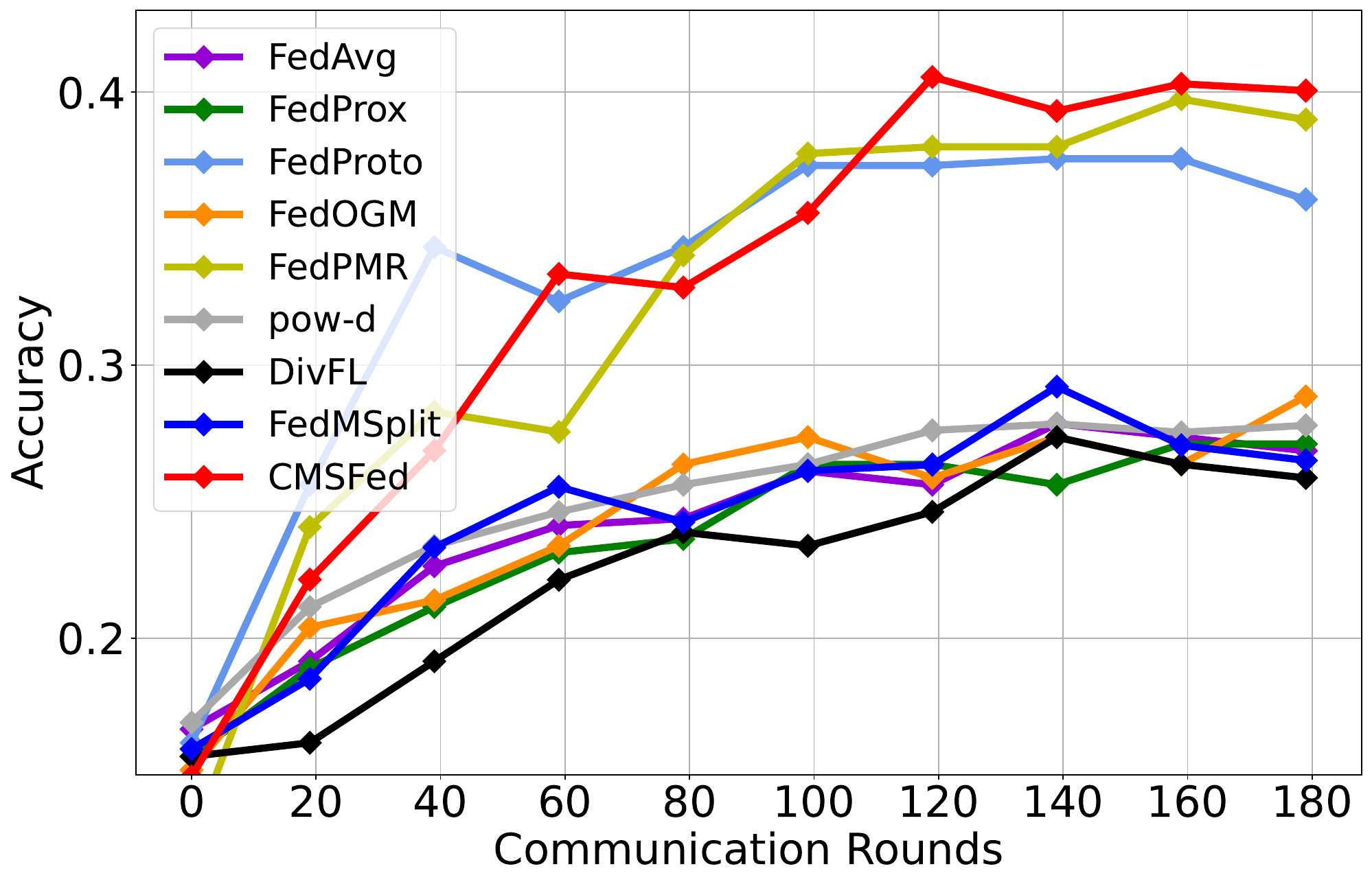}
        \caption{Visual in non-IID AVE}
        \label{fig:visual acc AVE}
    \end{subfigure}
    \caption{The performance of each modality compared with other baselines on CREMA-D and AVE under IID settings.}
    \label{fig:more modality performance IID}
\end{figure}

\begin{figure}[h]
    \centering
    \begin{subfigure}{0.45\linewidth}
        \includegraphics[width=0.9\linewidth]{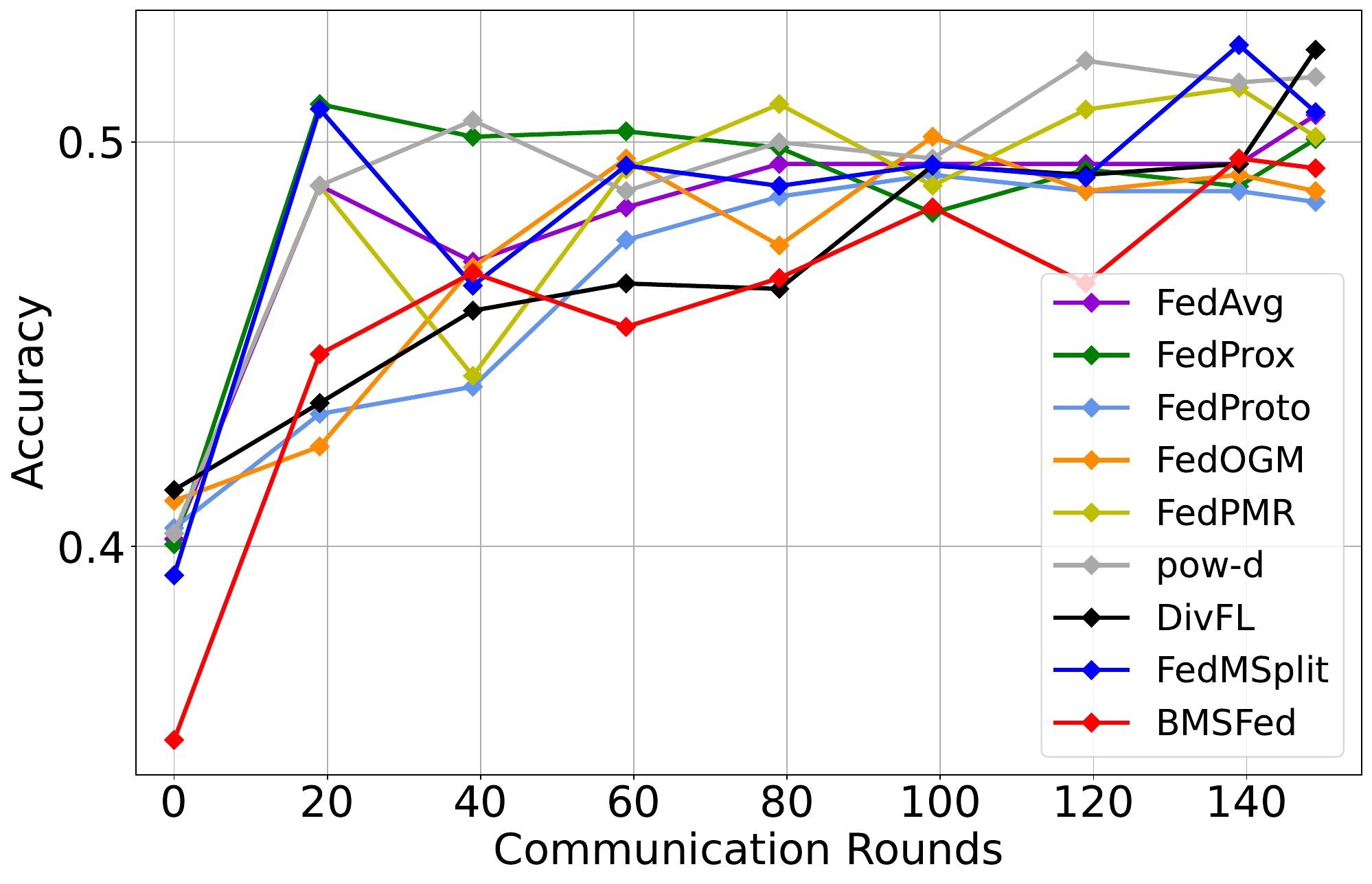}
        \caption{Audio in non-IID CREMA-D}
        \label{fig:audio acc CREMAD non-iid}
    \end{subfigure}
    \begin{subfigure}{0.45\linewidth}
        \includegraphics[width=0.9\linewidth]{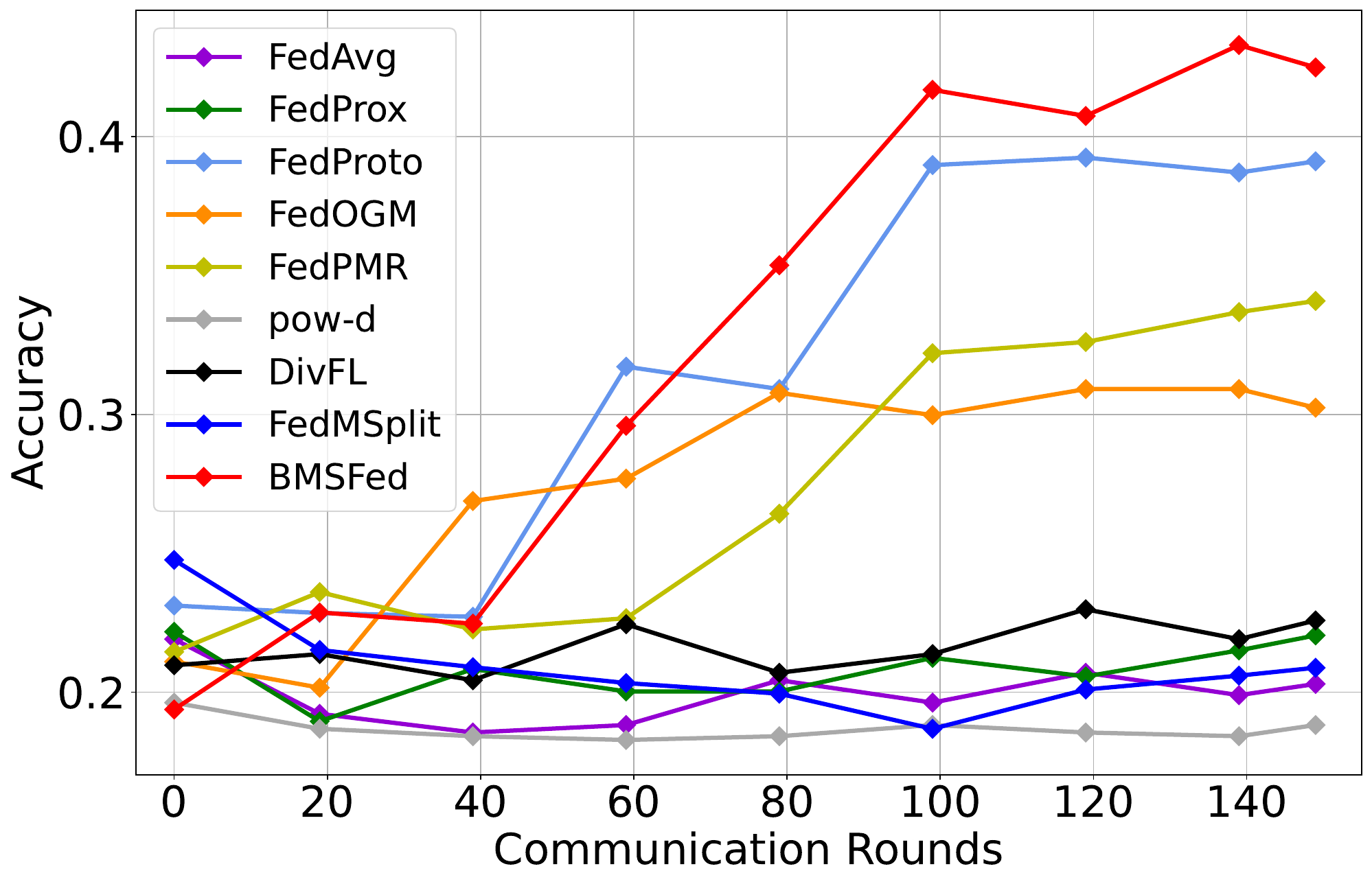}
        \caption{Visual in non-IID CREMA-D}
        \label{fig:visual acc CREMAD non-iid}
    \end{subfigure}
    
    \begin{subfigure}{0.45\linewidth}
        \includegraphics[width=0.9\linewidth]{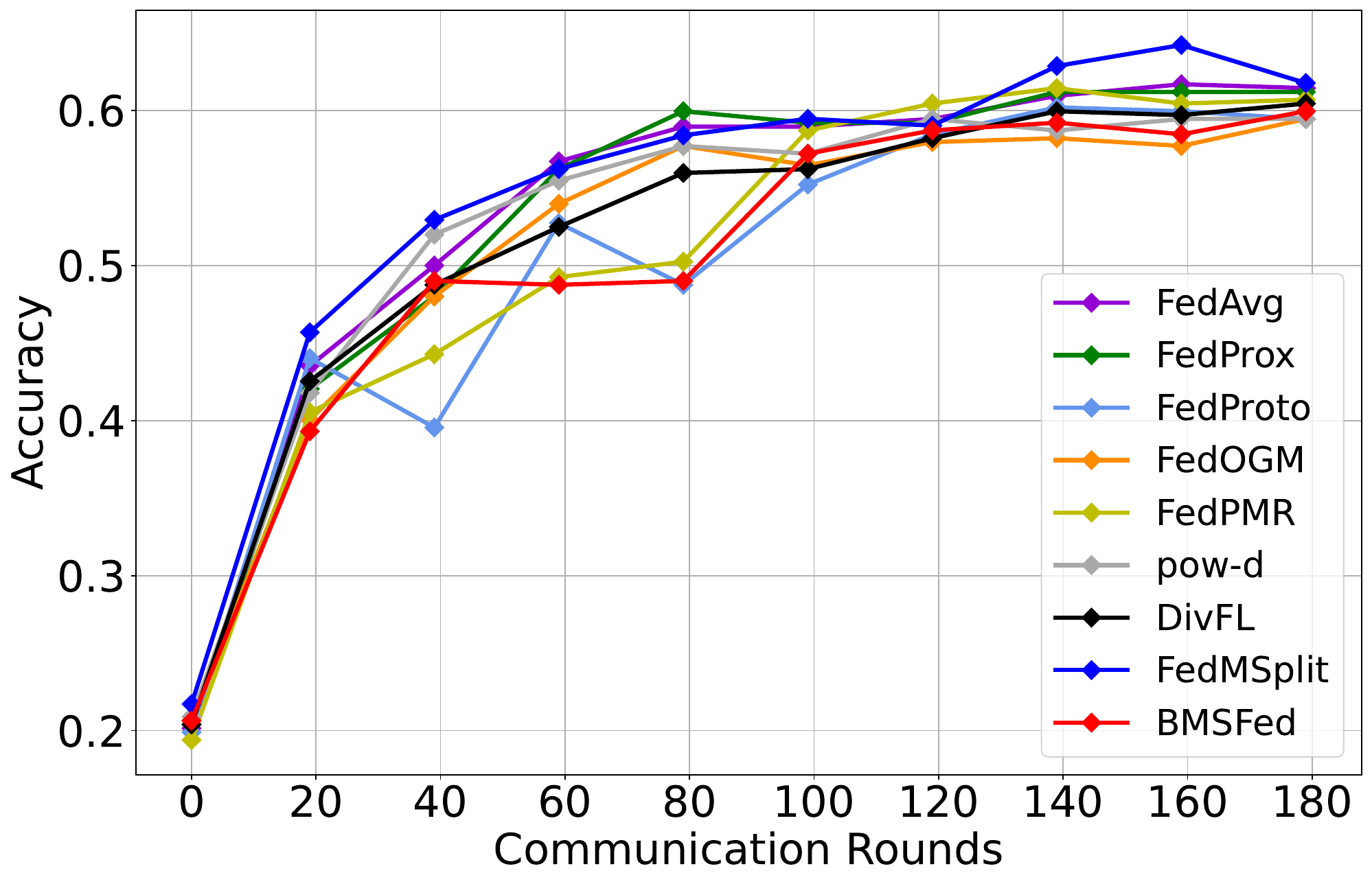}
        \caption{Audio in non-IID AVE}
        \label{fig:audio acc AVE}
    \end{subfigure}
    \begin{subfigure}{0.45\linewidth}
        \includegraphics[width=0.9\linewidth]{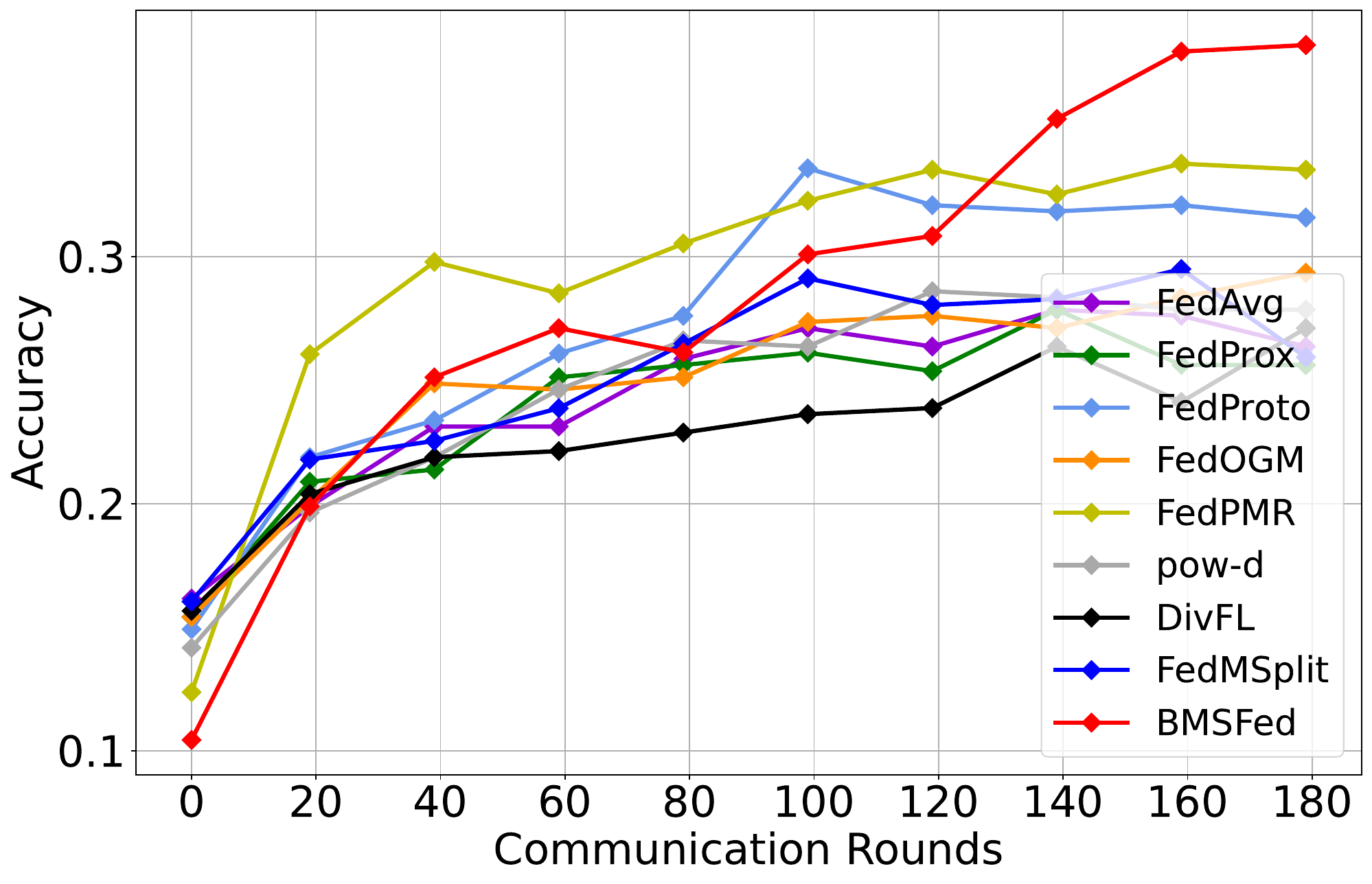}
        \caption{Visual in non-IID AVE}
        \label{fig:visual acc AVE}
    \end{subfigure}
    \caption{The performance of each modality compared with other baselines on CREMA-D and AVE under non-IID settings.}
    \label{fig:more modality performance non-IID}
\end{figure}

\textbf{The unimodal performance.} The accuracy curves of each modality of all baselines and our method on CREMA-D and AVE are shown Figs. \ref{fig:more modality performance IID} and \ref{fig:more modality performance non-IID}.

The uni-modal results under non-IID settings are consistent with the observations under IID settings.

\noindent\textbf{Comparison with more methods for modality imbalance.} Here are more results comparing our BMSFed with more baselines for modality imbalance. AGM \cite{li2023boosting}, G-blending \cite{wang2020makes} and Greedy \cite{wu2022characterizing} are all designed for modality imbalance problem in centralized scenario and we extend them to multi-modal FL settings. 
BMSFed still achieves the best performance on CREMA-D and AVE.

\begin{table}[h]
    \renewcommand\arraystretch{0.85}
	\centering
  \label{tab:more imbalance methods}
 \caption{Comparison with more modality imbalance methods. `C' and `A' denote CREMA-D and AVE respectively.}
    \setlength{\tabcolsep}{3pt}{
	\begin{tabular}{c|c|c|c|c}
        \hline
        ~ & C$_{IID}$ & C$_{non-IID}$ & A$_{IID}$ & A$_{non-IID}$ \\
        \hline
        AGM \cite{li2023boosting} & 55.1 & 52.1 & 63.4 & 60.1 \\
        G-blending \cite{wang2020makes} & 54.4 & 52.8 & 62.2 & 61.3 \\
        Greedy \cite{wu2022characterizing} & 54.0 & 53.9 & 62.9 & 60.7 \\
        \hline
        BMSFed & \textbf{64.5} & \textbf{61.6} & \textbf{64.7} & \textbf{62.1} \\
        \hline
	\end{tabular}
 }
\end{table}

\noindent\textbf{Comparison on Image-Text dataset with two sota FL methods for statistical heterogeneity.} We evaluate our method on the image-text dataset CrisisMMD \cite{abavisani2020multimodal} to show its effectiveness on text modality. Moreover, we choose two more SOTA FL methods FedNH \cite{dai2023tackling} and FedPAC \cite{xu2023personalized} for statistical heterogeneity FedNH and FedPAC for comparison. The results are shown in \cref{tab:text modality}, our method still achieves the best. 

\begin{table}[h]
\renewcommand\arraystretch{0.85}
\caption{Results compared with two sota FL methods on CREMA-D and an image-text dataset CrisisMMD.}
\label{tab:text modality}
\centering
\setlength{\tabcolsep}{3.5mm}{
\begin{tabular}{c|cc|cc}
\hline
          & \multicolumn{2}{c|}{CREMA-D} & \multicolumn{2}{c}{CrisisMMD \cite{abavisani2020multimodal}} \\ \hline
          & IID         & non-IID        & IID         & non-IID         \\ \hline
FedAvg    &  50.7           & 49.8               & 85.4            &  82.1               \\
FedNH     &  58.6           &  56.3              &  \underline{87.3}           & 85.8                \\
FedPAC    &  \underline{59.7}          &  \underline{56.4}              &  87.1           & \underline{86.5}                \\ \hline
FedPMR    &   55.5          &  55.1              &  86.6           & 86.0                \\
DivFL     &   51.7          &  50.8              &  85.8           & 83.5                \\
FedMSplit &   52.4          &  51.6              &  85.9           & 84.7                \\ \hline
BMSFed    & \textbf{64.5}            &  \textbf{61.6}         &   \textbf{88.7}          & \textbf{87.4}                \\ \hline
\end{tabular}
}
\end{table}

\noindent\textbf{More studies about claim and more settings.} We fix the selection scheme for each client but differs among clients. As shown in \cref{tab:claim}, this scheme is still better than FedAvg, proving ``uni-modal local training may contribute more to global model''. But they are worse than randomly modality selection, because some data is never selected. The results of CREMA-D with more clients with different $\alpha$ are in \cref{tab:alpha}.

\begin{table}[h]
\renewcommand\arraystretch{0.85}
\caption{More results proving ``uni-modal local training may contribute more to global model''.}
\label{tab:claim}
\centering
\setlength{\tabcolsep}{2.5mm}{
\begin{tabular}{c|ccccc}
\hline
    & FedAvg & -0.5 & -0.8 & -0.5-fix & -0.8-fix \\ \hline
A   & 51.2   & 50.5 & 48.1 & 49.6     & 45.2     \\ \hline
V   & 20.6   & 34.6 & 50.9 & 34.1     & 44.8     \\ \hline
A-V & 50.7   & 55.7 & 61.2 & 55.0     & 57.3     \\ \hline
\end{tabular}}
\end{table}

\begin{table}[h]
\renewcommand\arraystretch{0.85}
\caption{CREMA-D with more clients with different $\alpha$}
\label{tab:alpha}
\centering
\setlength{\tabcolsep}{1.8mm}{
\begin{tabular}{c|cccc}
\hline
       & $N$=20,$\alpha$=1 & $N$=20,$\alpha$=0.5 & $N$=40,$\alpha$=1 & $N$=40,$\alpha$=0.5 \\ \hline
FedAvg &  47.5        &  46.3         &   47.1                          &     42.9                          \\
BMSFed &  57.8        &  55.0         &   56.4                          &      53.3                         \\ \hline
\end{tabular}}
\end{table}

\end{document}